\pdfoutput=1
\documentclass{article}

\usepackage{algorithm}
\usepackage{algpseudocode}
\usepackage{newunicodechar}
\newunicodechar{≈}{\approx}

\PassOptionsToPackage{table}{xcolor}

\usepackage[authoryear, sort&compress, round]{natbib}
\usepackage[preprint,logo]{agibot_enerverse}

\usepackage[utf8]{inputenc}   
\usepackage[T1]{fontenc}      

\usepackage{booktabs}         
\usepackage{colortbl}         
\usepackage{multirow}         
\usepackage{longtable}        
\usepackage{soul}
\newcommand{\tablebodyfont}{\small}
\usepackage{amsmath}
\usepackage{amsfonts}         
\usepackage{nicefrac}         

\usepackage{graphicx}
\usepackage{caption}
\usepackage{subcaption}
\usepackage{wrapfig}
\usepackage{float}
\usepackage{placeins}          
\usepackage{afterpage}        
\usepackage{pifont}           

\usepackage{xcolor}

\usepackage{hyperref}         
\usepackage{url}
\usepackage{microtype}
\usepackage{xspace}           

\definecolor{BlueGreen}{rgb}{0.0, 0.55, 0.55}   
\definecolor{RedOrange}{rgb}{1.0, 0.27, 0.0}    

\newif\ifshowdraftids
\showdraftidsfalse
\ifshowdraftids
  \newcommand{\paraid}[1]{\mbox{\textcolor{gray}{\textsf{\scriptsize[#1]}}}\hspace{0.35em}}
  \newcommand{\sentid}[1]{\mbox{\textcolor{gray}{\textsuperscript{\textsf{\tiny S#1}}}}\hspace{0.12em}}
\else
  \newcommand{\paraid}[1]{}
  \newcommand{\sentid}[1]{}
\fi
\newcommand{\methodname}{KASO\xspace}                         
\newcommand{\methodnamegloss}{knowledge-aligned selective optimization\xspace} 
\newcommand{\gapname}{the validity gap\xspace}                
\newcommand{\coaename}{CoAE\xspace}                            
\newcommand{\svpname}{SVP\xspace}                              

\newcommand{\svpPretrainHours}{39{,}000} 
\newcommand{\idmPretrainHours}{32{,}000} 

\newcommand{\actionProbeTrainSamples}{400}
\newcommand{\actionProbeTestSamples}{100}
\newcommand{\coaeActionMAE}{0.01673}
\newcommand{\dcvaeActionMAE}{0.02573}
\newcommand{\dinoActionMAE}{0.01273}
\newcommand{\vjepaActionMAE}{0.01487}
\newcommand{\siglipActionMAE}{0.03138}
\newcommand{\coaeActionGapDino}{31}
\newcommand{\coaeActionGapVjepa}{13}
\newcommand{\dcvaeActionExcessOverCoAE}{54}
\newcommand{\siglipActionExcessOverCoAE}{88}

\newcommand{\instructionProbeEpisodes}{5{,}000}
\newcommand{\coaeInstructionAcc}{97.95}

\newcommand{\coaeInstructionError}{2.05}
\newcommand{\dcvaeInstructionError}{2.71}
\newcommand{\dinoInstructionError}{2.75}
\newcommand{\siglipInstructionError}{3.28}
\newcommand{\vjepaInstructionError}{4.15}
\newcommand{\coaeInstructionErrorReduction}{24}

\title{GE-Act 2.0: Pretraining and Scaling a World-Action Model for Robotic Manipulation}

\author{AgiBot Research Team\\[0.4em]
{\normalfont\normalsize \url{https://ge-act-v2.github.io/}}}
\hypersetup{pdfauthor={AgiBot Research Team}}

\begin{document}

\maketitle

\begin{abstract}

World--action models predict future states to guide robot actions, enabling
learning from both action-free video and action-labeled interaction. Yet most
inherit pretrained video generators, leaving pretraining and scaling of
world--action models underexplored. We introduce Genie Envisioner Act~2.0
(GE-Act~2.0), a world--action model in which every trainable generative and
action component is initialized from scratch on manipulation data. GE-Act~2.0
combines a control-oriented autoencoder (\coaename{}), a single-step visual
planner (\svpname{}), and an inverse dynamics model (IDM). \coaename{} retains
action- and instruction-relevant information under aggressive compression,
while \svpname{} produces a completed future state in one differentiable pass,
allowing visual planning and inverse dynamics to be pretrained separately on
complementary data. These components are then jointly trained using
\methodnamegloss{} (\methodname{}), which reduces mismatched supervision by
selecting only predicted future states judged behaviorally compatible with the
recorded action.
We evaluate pretrained checkpoints directly, without per-task fine-tuning, on
100 tasks across 20 manipulation skill groups, with held-out scenes,
backgrounds, lighting conditions, and object instances.
Scaling co-training data from 300 to 30{,}000 hours raises success
from 17.1\% to 44.1\% on G1-OP and from 13.4\% to 31.1\% on G2-90D. Despite
comprising less than 2\% of the co-training data, G2-90D improves by 17.7
points, suggesting useful cross-embodiment transfer from the broader corpus.
The gains span 19/20 and 18/20 skill groups, respectively, and skill-specific
coverage strongly correlates with zero-shot out-of-distribution (OOD) success (Pearson $r=0.80$;
Spearman $\rho=0.85$). Under the same zero-shot OOD protocol, the model grounds object,
color, shape, and position references in at least 90\% of trials; qualitative stress
tests further show that it follows explicit instructions even when they
conflict with an already-committed behavior or a conventional scene
association.

\end{abstract}

\afterpage{%
\begin{figure}[t!]
\centering
\includegraphics[width=\linewidth]{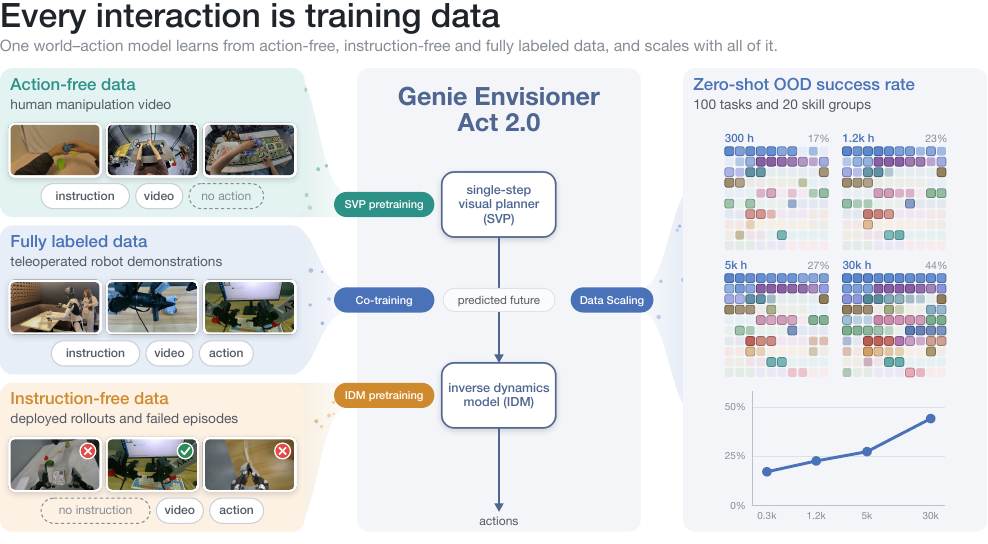}
\caption{GE-Act~2.0 pretrains its visual planner on video without action labels
and its inverse dynamics model on trajectories without language instructions,
then co-trains the full system on fully labeled robot demonstrations.  Scaling
co-training data from 300 to 30{,}000 hours improves zero-shot OOD success
across a 100-task real-robot suite.  Each tile in the right panel is one task,
coloured by skill group and shaded by success rate; the curve is the mean over
all tasks.}
\label{fig:teaser}
\end{figure}%
}

\section{Introduction}
\label{sec:intro}
Large-scale pretraining has driven successive breakthroughs in language, multimodal understanding, and visual generation. In robot manipulation, vision--language--action (VLA) policies map observations and instructions directly to motor commands at scale~\citep{brohan2022rt,black2024pi0,intelligence2025pi05}, but their direct action objective does not explicitly require modeling the physical dynamics underlying interaction. World-action models (WAMs) are emerging as a new paradigm: they predict how a scene may unfold and use that prediction to support action generation~\citep{wang2026wamsurvey,huang2025enerverse,liao2025genie,zhang2026lingbot,zhang2026lingbotva2,amap2026abotm05,ye2026dreamzero}. Besides making the model's visual predictions inspectable, this formulation creates two routes to broader pretraining data. Visual-generation pretraining can capture physical and semantic structure from action-free video at a scale that robot demonstrations alone cannot provide; an inverse dynamics model (IDM) can potentially learn from robot trajectories that lack instruction or success annotations, including failed attempts and deployment rollouts.

However, existing WAM research typically builds on pretrained video generators~\citep{wan2025,nvidia2025cosmos} and devotes substantial effort to coupling them with action models~\citep{wang2026wamsurvey,liao2025genie,Team2026MotuBrainAA,wu2024gr1,ye2026dreamzero}, rather than to pretraining and scaling the WAM as a whole. This leaves two system-level questions unresolved: how to pretrain visual generation and inverse dynamics on their respective data before connecting them, and how WAM capability scales with manipulation data when all trainable visual-generation and action components are initialized from scratch.

Genie Envisioner Act~2.0 (GE-Act~2.0) therefore adopts a simple decoupled two-stage architecture and focuses on three design problems for scalable WAM pretraining: constructing the latent space, predicting actionable future visual states in that space, and aligning those states with recorded action supervision.

\begin{description}[leftmargin=1.8em, itemsep=0.35em, topsep=0.45em, font=\normalfont\bfseries]
  \item[Representation:] A latent space for robot manipulation must make future prediction efficient while retaining the visual information needed for action prediction. Most systems inherit a space from a standard video autoencoder~\citep{wan2025,nvidia2025cosmos}, optimized for reconstruction fidelity rather than control. We therefore introduce a control-oriented autoencoder (CoAE) that learns an aggressively compressed latent space through reconstruction and multi-teacher feature alignment, preserving the information needed for both visual prediction and action recovery (Section~\ref{sec:representation}).
  \item[Generation:] Visual generation and inverse dynamics can exploit different types of data. Visual-generation pretraining can use action-free video, whereas IDM pretraining can use instruction-free robot trajectories such as failed attempts and deployment rollouts, which are difficult to exploit under conventional policy-training objectives. However, most prior WAMs do not support standalone IDM pretraining on this broader interaction data~\citep{hu2025vpp,ye2026dreamzero,liao2025genie}; in Section~\ref{sec:why_onestep}, we trace this limitation to the cost of multi-step generation. We therefore introduce a single-step visual planner (SVP), which produces a complete future in one differentiable flow-generator pass and allows the SVP and IDM to be pretrained separately on complementary data before being connected through end-to-end co-training (Section~\ref{sec:generation}).
  \item[Alignment:] Even when the robot sees the same scene and receives the same instruction, there may be several correct ways to complete the task. In imitation learning, a recorded demonstration contains only one of these valid behaviors, while an independently generated future may depict another. Pairing that future with the recorded action gives the IDM mismatched visual and action supervision, creating what we term \textbf{\gapname{}}. Repeated mismatches can erase distinct action modes and reduce the action diversity available for later exploration. \methodname{} (\methodnamegloss{}) samples multiple futures and co-trains only on candidates that the active IDM judges compatible with the recorded mode, preserving broader action-space coverage for subsequent post-training methods such as reinforcement learning (Section~\ref{sec:validity}).
\end{description}

Evaluating what pretraining contributes is therefore the central empirical question for this design. Many evaluations of WAMs and VLA policies use downstream supervised fine-tuning (SFT) on the target tasks or embodiments~\citep{zhang2026lingbot,zhang2026lingbotva2,gigaworldpolicy2026,zhou2026tau0wm,wu2024gr1,cheang2024gr2,kim2024openvla,black2024pi0,brohan2022rt}. Some further evaluate under the same lighting, backgrounds, and object sets used for SFT, making the test effectively in-domain. These protocols characterize the final policy, but entangle capability acquired during pretraining with capability added downstream. We instead evaluate GE-Act~2.0 directly under out-of-distribution (OOD) conditions, without task- or embodiment-specific SFT, and keep the protocol fixed across scaling checkpoints. This design measures the capability already present after pretraining and how it changes with training data.

\paragraph{Contributions.}
\begin{itemize}[leftmargin=1.6em, itemsep=0.2em, topsep=0.3em]
  \item We introduce GE-Act~2.0, a world--action model designed for pretraining
  from scratch on manipulation data. Its control-oriented autoencoder provides
  a compact representation for visual prediction and action recovery, while
  its single-step visual planner allows visual planning and inverse dynamics
  to be pretrained separately on complementary data and subsequently connected
  through end-to-end co-training.

  \item We identify \gapname{}, a supervision mismatch that arises when a
  predicted future depicts a behavior different from the recorded action, and
  propose \methodname{} to co-train GE-Act~2.0 using action-compatible predicted
  futures.

  \item We demonstrate systematic scaling of GE-Act~2.0's pretrained zero-shot
  OOD capability: increasing manipulation data produces broad gains across 100
  real-robot tasks and two embodiments without task-specific fine-tuning. The
  results further reveal a strong relationship between skill coverage and
  success, as well as useful transfer to a sparsely represented embodiment.
  Under the same OOD protocol, complementary language experiments demonstrate
  fine-grained referential grounding and instruction following when explicit
  commands conflict with already-committed behaviors or conventional scene
  associations.
\end{itemize}

\section{Related Work}
\label{sec:related}

\subsection{Generalist Policies and World-Action Models}
\label{sec:related:wam}

Generalist robot policies scale language-conditioned control across broad task
collections~\citep{brohan2022rt,black2024pi0,intelligence2025pi05,team2024octo,kim2024openvla,liu2025geniereasoner,generalist2025gen0,generalist2026gen1}.
World-action models connect visual prediction to control through several
interfaces~\citep{wang2026wamsurvey}. Decoupled two-stage designs first generate
future visual states and then infer actions with a separate action
model~\citep{jang2025dreamgen};
joint designs predict video and actions in a shared
backbone~\citep{wu2024gr1,cheang2024gr2,Team2026MotuBrainAA,ye2026dreamzero};
and feature-coupled designs decode actions from internal representations of a
generative backbone~\citep{liao2025genie,hu2025vpp}.
GE-Act~2.0 uses an explicit completed future as the interface to a separately
pretrained IDM. This differs from the parallel action branch of GE-Act~1.0 and
allows the generator and IDM to pretrain on distinct data before end-to-end
co-training.

\subsection{Representations for Robot Video and Control}
\label{sec:related:representation}

Most latent video generators operate in autoencoder spaces developed for content reconstruction~\citep{wan2025,nvidia2025cosmos}. Robot learning has also explored representations designed around motion or action: Genie discovers discrete latent actions from video~\citep{bruce2024genie}; LAPA and Moto connect learned motion tokens to robot actions~\citep{ye2025lapa,chen2025moto}; and IGOR compresses visual change into a shared latent action space~\citep{chen2024igor}. Other work uses predictive visual features directly for control, including video-pretrained backbones~\citep{wu2024gr1,cheang2024gr2}, diffusion features~\citep{hu2025vpp}, and point trajectories~\citep{wen2024atm}. In image generation, representation alignment methods such as REPA and VA-VAE show that matching a tokenizer's latents or a generator's intermediate features to frozen semantic encoders can improve generation~\citep{yu2025repa,yao2025vavae}. \coaename{} addresses both concerns: it provides a compact latent space with a reconstruction decoder for generation, is aligned to multiple pretrained visual teachers, and is evaluated by how well a fixed-capacity action probe can recover actions from its encoded video windows.

\subsection{Fast Video Generation for World Models}
\label{sec:related:generation}

Robot world models often adapt a video generator pretrained on natural video, then retain multi-step diffusion or compress the sampler through distillation~\citep{jang2025dreamgen,guo2025ctrl,huang2025enerverse,jiang2025enerverseac,qiu2026gesim2,zhu2024irasim,zhang2026lingbot,zhang2026lingbotva2,amap2026abotm05,nvidia2025cosmos}. One- and few-step generation has a broader lineage in images: progressive and distribution-matching distillation compress a pretrained teacher~\citep{salimans2022progressive,yin2024dmd,yin2024improved}; consistency models enforce agreement along the probability-flow trajectory~\citep{song2023consistency,luo2023latentconsistency,kim2024consistencytrajectory}; and rectified or shortcut flows learn paths intended for small integration budgets~\citep{lipman2023flowmatching,liu2023rectifiedflow,liu2024instaflow,frans2025shortcut}. MeanFlow and improved MeanFlow instead train an average-velocity field that can traverse the complete noise interval in one network forward pass without first distilling a multi-step teacher~\citep{geng2025meanflow,geng2026improvedmeanflows}. Seaweed-APT demonstrates one-step natural-video generation through adversarial post-training~\citep{lin2025seaweedapt}. The closest methodological starting points for our single-step visual planner are MeanFlow and improved MeanFlow; our focus is their conditional extension to multi-view future visual states and their use as a differentiable train--deployment interface for an action model.

\subsection{Connecting Generated Futures to Actions}
\label{sec:related:alignment}

World models used only for planning, latent imagination, or representation extraction need not expose an action model to generated future visual-state samples~\citep{Hafner2018LearningLD,Hafner2019DreamTC,Wu2022DayDreamerWM,hu2025vpp,qiu2026gesim2}. World-action training creates a stronger coupling. Teacher forcing trains the action model on recorded future visual states, leaving a generated-versus-recorded input gap at deployment, as in LingBot-VA~\citep{zhang2026lingbot,zhang2026lingbotva2}; other systems feed generated future visual states but detach the sampling path or rely on robustness to absorb generation errors~\citep{amap2026abotm05}. These approaches primarily address whether generated video is used and whether action gradients reach the video generator. \methodname{} targets a different mismatch: even a plausible visual prediction may represent a different outcome mode from the recorded action used as its label. It samples several candidates, ranks them with the active IDM against the recorded future visual states, and applies generated-visual-state action supervision only to the compatible candidates (Section~\ref{sec:validity}). We also note that a similar best-of-$K$ selection rule has appeared
independently in concurrent work on general generative
modeling~\citep{gladstone2026explorative}, where training on the best of $K$
candidate matches lets predictions commit to modes rather than blur them;
\methodname{} instead judges compatibility in action space with the active IDM
and uses the selection to align the visual planner with the IDM.

\subsection{Evaluating Pretrained Robot Models}
\label{sec:related:eval}

Robot models are commonly evaluated after task-specific fine-tuning~\citep{liao2025genie,zhang2026lingbot,zhang2026lingbotva2,gigaworldpolicy2026,zhou2026tau0wm,amap2026abotm05,wu2024gr1,cheang2024gr2}, while fewer systems report zero-shot results~\citep{zhou2026tau0wm,intelligence2025pi05,generalist2025gen0,generalist2026gen1}. Fine-tuned performance is important, but it combines the quality of the pretrained model with the amount and coverage of adaptation data. We therefore report no-fine-tuning evaluation as the primary measure of pretrained capability and state which visual factors are excluded and which forms of task adaptation are absent (Section~\ref{sec:pretrained_eval}). Simulation and world-model benchmarks~\citep{liu2023libero,james2020rlbench,shang2026worldarena} provide horizontal reference points rather than direct evidence for the capability acquired by our manipulation-video pretraining.

\section{GE-Act 2.0}
\label{sec:method}

\raggedbottom


\subsection{System Overview}
\label{sec:overview}

\begin{figure}[!t]
\centering
\includegraphics[width=\linewidth]{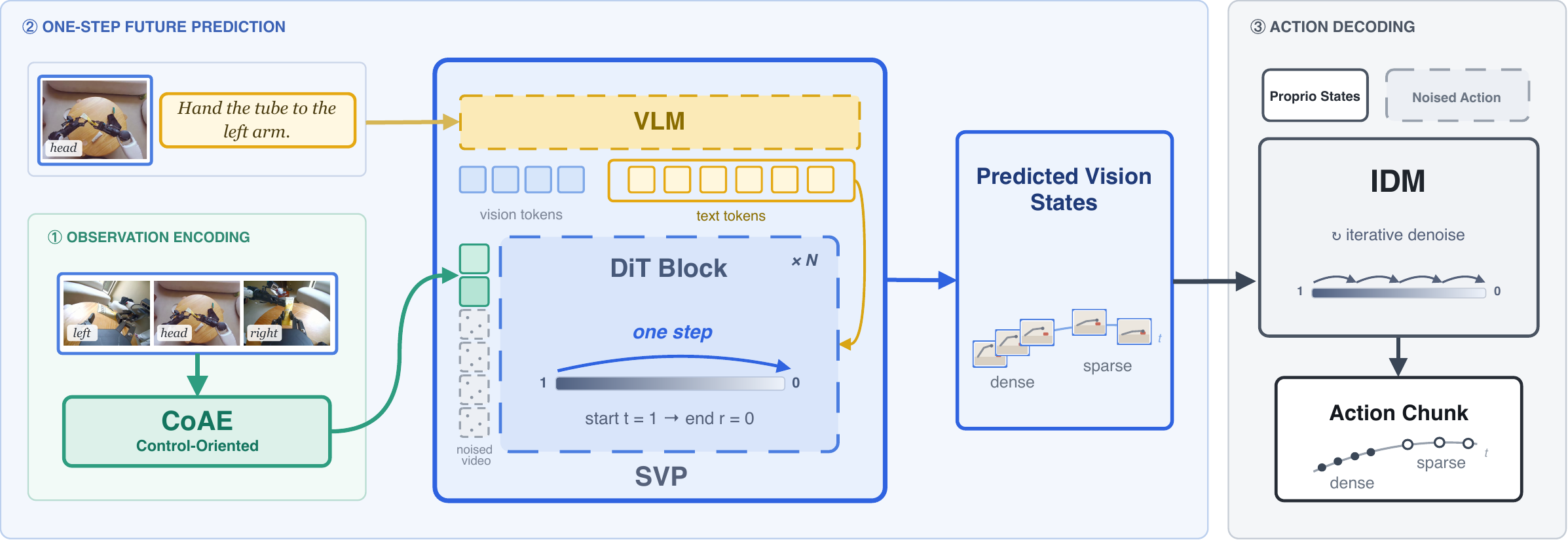}
\caption{GE-Act~2.0 system architecture. The \svpname{} grounds the instruction
in the current observation and fully denoises multi-view future visual latents in
a single flow-generator pass; the IDM combines current and predicted latents
with proprioception to produce dense and sparse action sequences.}
\label{fig:system}
\end{figure}

GE-Act~2.0 is a world-action model designed for from-scratch pretraining and
scaling with heterogeneous video and robot-interaction data. As summarized in
Figure~\ref{fig:system}, it comprises a control-oriented autoencoder
(\coaename), a single-step visual planner (\svpname), and an inverse dynamics
model (IDM). The \svpname serves as the system's generative world model. We train
\coaename{}, and we randomly initialize and pretrain every other trainable
component from scratch: the flow generator in the \svpname{}, and the IDM.

\coaename{} first encodes the current multi-view observations into compact
visual latents. Conditioned on these latents and the instruction, the
\svpname{} grounds the instruction in the current scene and fully denoises
dense and sparse future visual latents in a single flow-generator pass. The
fully denoised future
visual states are passed to the IDM, which combines current and predicted
visual latents with proprioception to predict dense and sparse sequences of
action; the controller executes the dense chunk, while
the sparse sequence provides auxiliary far-horizon targets during training.

The remainder of this section develops these components and their interaction.
Section~\ref{sec:representation} introduces \coaename{} and its compact visual
space, Section~\ref{sec:generation} develops the \svpname{}, and
Section~\ref{sec:validity} connects generated visual futures to action learning
through the IDM and \methodname{}.

\subsection{A Control-Oriented Autoencoder for World--Action Models}
\label{sec:representation}

A latent space for a world-action model determines how efficiently and
effectively a future can be predicted, as well as what information the action
model can recover from that prediction.  Yet most existing WAMs inherit this
space from a pretrained video
generator~\citep{jang2025dreamgen,guo2025ctrl,zhang2026lingbot,zhang2026lingbotva2,amap2026abotm05}.
This inheritance creates two mismatches.  First, the modest spatial
compression of standard video autoencoders leaves dense latent grids,
imposing a large token burden on every future prediction inside the control
loop.  Second, these autoencoders are trained primarily to reconstruct generic
images and videos, rather than explicitly designed to support downstream
visual planning and action prediction.

Our control-oriented autoencoder (\coaename{}) addresses both limitations by
learning an aggressively compressed,
control-oriented latent space from manipulation video.  Its training uses pixel
reconstruction together with alignment to complementary visual teachers, and its
latents are consumed directly by both the \svpname{} and the inverse-dynamics
model.  We next describe its training objective and evaluate the resulting
representation through action recovery and instruction grounding.

\raggedbottom
\subsubsection{Beyond Reconstruction with Multi-Teacher Alignment}
\label{sec:coae_learning}

\coaename{} is a framewise 2D autoencoder designed to produce a short spatial
token sequence.  Given an input frame $o$, its encoder produces
$z=\mathcal{E}(o)$.  Whereas commonly used video autoencoders downsample each
spatial axis by $8\times$ or $16\times$~\citep{wan2025,nvidia2025cosmos},
\coaename{} uses a $64\times$ spatial downsampling factor.  At the
$256{\times}384$ frame resolution used by the \svpname{}
(Section~\ref{sec:generation}), this maps each input frame to a
$4{\times}6$ grid of 512-channel latents, corresponding to 24 tokens per
frame.  We initialize \coaename{} from the 128-channel
DC-AE~\citep{chen2025dcae} by transferring its compatible weights, while
expanding the latent width to 512 channels and randomly initializing the newly
introduced parameters.

\begin{figure}[H]
\centering
\includegraphics[width=\linewidth]{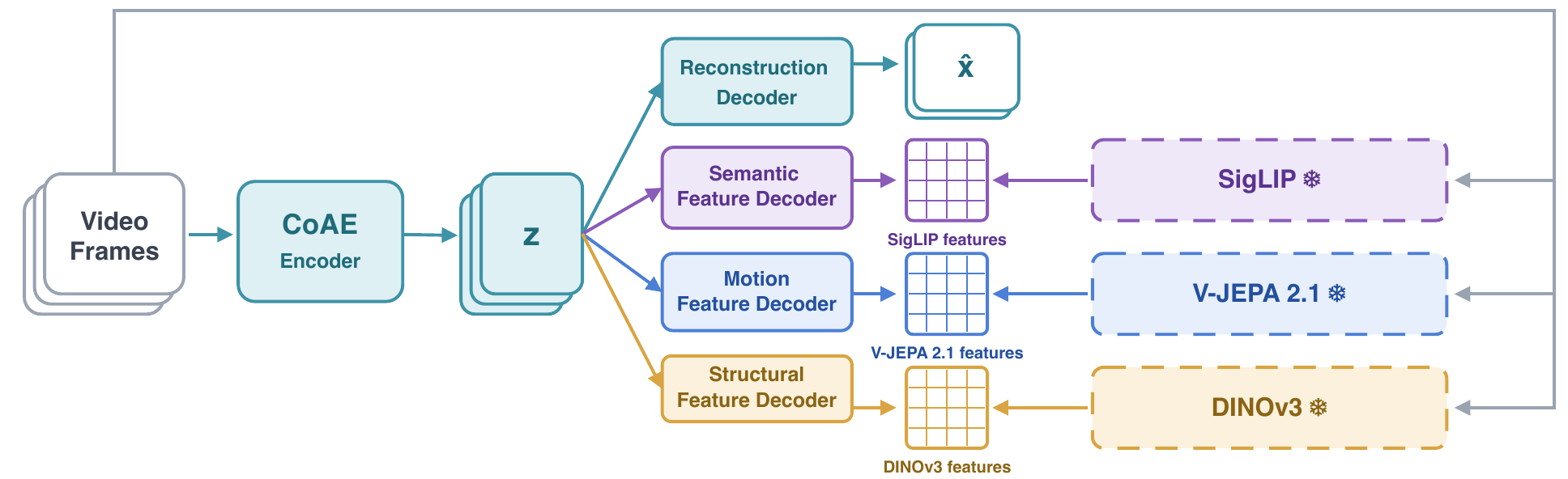}
\caption{\coaename{} architecture and training objectives.  At the
$256{\times}384$ frame resolution used by the \svpname{}, the encoder maps one
input frame to a $4{\times}6$ grid of 512-channel latents, giving 24 tokens per
frame.  A reconstruction decoder provides pixel-level supervision, while
three alignment heads match language-aligned, spatiotemporal, and dense visual
features from frozen visual teachers.}
\label{fig:coae_architecture_comparison}
\end{figure}

During autoencoder training, a decoder reconstructs the input as
$\hat{o}=\mathcal{D}(z)$.  We supervise this reconstruction with pixel,
perceptual~(LPIPS), and adversarial losses:
\begin{equation}\label{eq:coae_rec_comparison}
\mathcal{L}_{\mathrm{rec}} =
\bigl\|\hat{o}-o\bigr\|_2^2
+ \lambda_{\mathrm{perc}}\,
  \mathcal{L}_{\mathrm{perc}}\bigl(\hat{o},o\bigr)
+ \lambda_{\mathrm{gan}}\,
  \mathcal{L}_{\mathrm{gan}}\bigl(\hat{o}\bigr).
\end{equation}
Reconstruction alone does not directly supervise the semantic and
spatiotemporal information relevant to action control.  Motivated by evidence that representation
objectives can improve generative models trained over learned visual
tokens~\citep{yao2025vavae,yao2025vtp}, we align $z$ with three frozen visual teachers.
Separate alignment heads $g_k$ map $z$ toward the final-layer representations
$G_k(o)$ of the SigLIP~2 vision encoder used by Qwen3.5, V-JEPA~2.1, and
DINOv3~\citep{zhai2023siglip,tschannen2025siglip2,murlabadia2026vjepa21,simeoni2025dinov3}.
These teachers provide complementary language-aligned semantics,
spatiotemporal structure, and dense visual features, respectively.  The
alignment objective is
\begin{equation}\label{eq:coae_align_comparison}
\mathcal{L}_{\mathrm{align}} =
\sum_{k=1}^{3}
\lambda_k
\left(
1-\cos\bigl(g_k(z),G_k(o)\bigr)
\right).
\end{equation}
The complete training objective is
$\mathcal{L}_{\mathrm{CoAE}}
=\mathcal{L}_{\mathrm{rec}}+\mathcal{L}_{\mathrm{align}}$.
Figure~\ref{fig:coae_architecture_comparison}
summarizes the architecture and training objectives.

\flushbottom

\subsubsection{Control and Instruction Grounding under Extreme Compression}
\label{sec:coae_probes_comparison}

We evaluate the compressed representation through action recovery and
instruction grounding, comparing \coaename{} with its three visual teachers and
DC-AE~\citep{chen2025dcae}, a deep compression autoencoder with the same $64{\times}$ spatial downsampling factor as
\coaename{}.

\begingroup
\textit{Action recovery.}\quad
Action recovery is our primary probe because \coaename{} is the latent
interface through which predicted visual changes inform the IDM.  Unlike
reconstruction metrics, which assess appearance fidelity, inverse-dynamics
probing directly tests whether the compressed latent preserves the visual state
changes needed to infer executed robot actions.  We freeze each encoder and
train the same two-layer inverse-dynamics probe
(Section~\ref{sec:idm}) on a set of \actionProbeTrainSamples{}
samples from a visually cluttered shelf task that requires the robot to
categorize and organize many items.  Each sample pairs dense and sparse frames
with the corresponding recorded actions.  We report action mean absolute error
(MAE) on \actionProbeTestSamples{} held-out samples from the same task.  All
representations use the same train--test split and optimization protocol.
Because every encoder is frozen and evaluated with the same two-layer probe,
differences in MAE indicate how directly each representation exposes
control-relevant information.  As shown in
Table~\ref{tab:coae_probe_comparison}, DINOv3 and
V-JEPA~2.1 obtain the lowest error.  \coaename{} follows at \coaeActionMAE{},
trailing them by only \coaeActionGapVjepa--\coaeActionGapDino\% while using
one-sixteenth as many tokens per frame.  In contrast, the errors of DC-AE and
SigLIP are \dcvaeActionExcessOverCoAE\% and \siglipActionExcessOverCoAE\% higher
than that of \coaename{}, respectively.

\begin{table}[H]
\captionsetup{skip=8pt}
\caption{Controlled probes of five frozen visual representations.  Spatial
downsampling is reported as the height $\times$ width factor relative to the
input; for example, $64{\times}64$ means that one latent location spans a
$64{\times}64$-pixel input region.  Channels gives the latent feature
dimension.  Lower is better for both probe metrics.}
\label{tab:coae_probe_comparison}
\centering
\tablebodyfont
\setlength{\tabcolsep}{5pt}
\begin{tabular}{lccccc}
\toprule
Representation & Spatial downsampling & Channels & Pixel decoder & Action MAE $\downarrow$
  & Caption-match error (\%) $\downarrow$ \\
\midrule
\rowcolor{RedOrange!8}
\textbf{\coaename} & $64{\times}64$ & 512 & \ding{51}
  & \coaeActionMAE & \textbf{\coaeInstructionError} \\
DC-AE & $64{\times}64$ & 128 & \ding{51}
  & \dcvaeActionMAE & \dcvaeInstructionError \\
DINOv3 & $16{\times}16$ & 1024 & \ding{55}
  & \textbf{\dinoActionMAE} & \dinoInstructionError \\
V-JEPA~2.1 & $16{\times}16$ & 1024 & \ding{55}
  & \vjepaActionMAE & \vjepaInstructionError \\
SigLIP & $32{\times}32$ & 2048 & \ding{55}
  & \siglipActionMAE & \siglipInstructionError \\
\bottomrule
\end{tabular}
\end{table}

\textit{Instruction grounding.}\quad
As a complementary probe, instruction grounding tests whether the compact
latent preserves how the current scene relates to the commanded behavior.  We
construct and evaluate the caption-matching probe using
\instructionProbeEpisodes{} GenieSim-Instruction episodes.  The episodes
are distributed uniformly across all ten benchmark tasks and instruction
categories, providing balanced coverage of color, number, shape, size, object
type, specific-object reference, common-sense reference, logical composition,
and object straightening.  The complete task list appears in
Appendix~\ref{app:simulation:geniesim} and
Table~\ref{tab:geniesim_instruction}.  A positive pair contains an observation
and its original instruction, whereas a negative pair uses an instruction
sampled from another episode; the two classes are sampled equally.  For each
frozen encoder, an eight-layer network following our video-generator
architecture (Section~\ref{sec:generation}) fuses the latent tokens and
candidate instruction while retaining VLM conditioning; a two-layer MLP
classifies the average-pooled hidden states.  We optimize the probe with binary
cross-entropy and report classification accuracy under the same protocol for
all representations.
\coaename{} achieves
\coaeInstructionAcc\% accuracy, ahead of every teacher and the
reconstruction-only baseline.  Its \coaeInstructionError\% error rate is
\coaeInstructionErrorReduction\% lower than the next-best DC-AE and roughly
half the \vjepaInstructionError\% error of V-JEPA~2.1.

\par\vspace{0.5\baselineskip}
\noindent
\begin{minipage}[t]{0.395\linewidth}
\vspace{0pt}
\raggedright
Figure~\ref{fig:coae_probe_comparison} summarizes the central result:
our \coaename{} occupies a favorable compactness--informativeness operating point.
On action recovery, it remains within
\coaeActionGapVjepa--\coaeActionGapDino\% of DINOv3 and V-JEPA~2.1 while using
one-sixteenth as many tokens per frame.  Despite using the same $64{\times}$
downsampling along each spatial axis, DC-AE performs worse on both probes,
and \coaename{} has the lowest instruction-matching error of all five, showing
that compression rate alone does not explain its result.  \coaename{} and
DC-AE are also the only two representations that keep a pixel decoder, so this
operating point costs no decodability.  We attribute this to
manipulation-domain training combined with multi-teacher alignment, which
supplies language-aligned semantics, spatiotemporal structure, and dense visual
features.
\end{minipage}\hfill
\begin{minipage}[t]{0.575\linewidth}
\vspace{0pt}
\centering
\includegraphics[width=\linewidth,trim=8bp 9bp 7bp 7bp,clip]{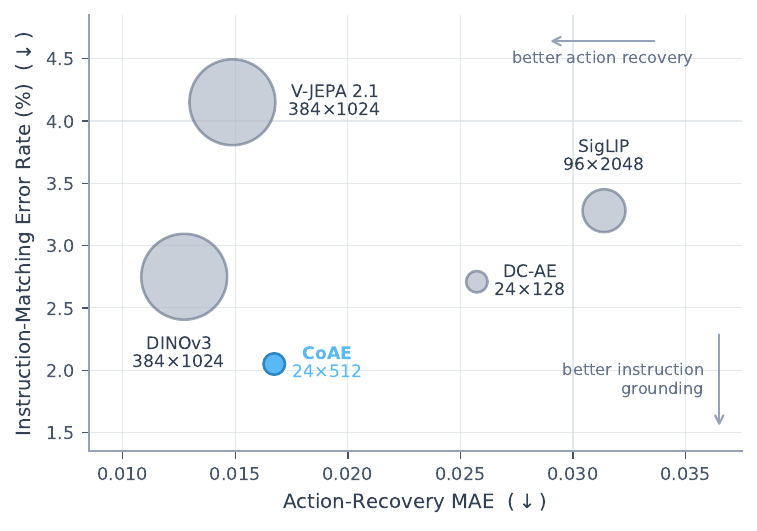}
\captionsetup{font=footnotesize,skip=5pt,hypcap=false}
\captionof{figure}{Action-recovery and instruction-grounding probes for five
frozen visual representations.  Lower is better on both axes.  For each
$256{\times}384$ input frame, labels report tokens per frame $\times$ channel
dimension; marker area scales with tokens per frame.}
\label{fig:coae_probe_comparison}
\end{minipage}
\par\vspace{0.75\baselineskip}
\endgroup

\subsection{A Single-Step Visual Planner as a World--Action Interface}
\label{sec:generation}

The single-step visual planner (\svpname{}) is the visual-generation component
of GE-Act~2.0 and provides an
explicit future-visual interface to a separately pretrained IDM. It comprises a
multimodal understanding module and a conditional flow generator. The flow
generator adopts the multi-view diffusion transformer (DiT) pattern of
GE-Act~1.0~\citep{liao2025genie,hacohen2024ltxvideo}, interleaving per-view space--time processing
with periodic cross-view attention. Given a current observation and instruction,
the multimodal understanding module grounds the instruction in the observed
scene, and the flow generator fully denoises the corresponding future visual
latents in a single flow-generator pass. The generated future is then passed to
the IDM during co-training and deployment. To our knowledge, our \svpname{} is the
first native single-step robot world model pretrained from random
initialization on manipulation video.

\subsubsection{One-Step Generation Unlocks Underutilized Data}
\label{sec:why_onestep}

High-quality robot demonstrations remain scarce, making data a key bottleneck
to scaling robot learning. Consequently, whether a system can exploit diverse
data beyond successful, instruction-annotated trajectories is a central design
question. An IDM can learn from robot trajectories with aligned observations
and actions, including instruction-free play, failed attempts, and deployment
rollouts, without requiring language or success annotations. Such data are
difficult to exploit efficiently under the imitation-learning objectives used
by VLAs and prior WAMs.

However, using these broader data within a WAM requires more than pretraining
the IDM separately: the pretrained IDM must subsequently be connected to the
\svpname{} for end-to-end co-training. During separate pretraining, the IDM
receives future observations from recorded trajectories. During co-training,
the same input must instead come from the visual planner, and the action loss must
propagate through the generated future to update both components. With a
conventional multi-step video generator, however, a complete future is obtained
only after $K$ sequential denoising steps. Propagating action gradients back to
the generator therefore requires retaining and differentiating through the
entire $K$-step generation process, substantially increasing computation and
memory. Systems such as VPP, DreamZero, and GE-Act
1.0~\citep{hu2025vpp,ye2026dreamzero,liao2025genie} therefore avoid this cost by
predicting actions from intermediate states of the generator. Their action
components consequently remain tied to the visual generator and do not provide
a separate training path for the broader interaction data described above.

Our single-step visual planner resolves this limitation by producing a complete future in one
differentiable generator pass. The IDM can therefore first be pretrained on
recorded robot trajectories and then connected to the \svpname{} for
end-to-end co-training. During co-training, the IDM predicts actions from
futures generated by the \svpname{}, and the action loss updates both the IDM and
the \svpname{}. At deployment, the IDM continues to predict actions from futures
generated by the \svpname{}.

\subsubsection{Multi-Scale Visual Prediction}
\label{sec:mstp}

A useful visual prediction must explain both how the scene changes next and
where the task is headed. Dense near-term targets train the \svpname{} to resolve
fine motion and interaction dynamics, while sparse far-horizon targets require
it to preserve task intent and model longer-term scene evolution. Training at
both scales encourages the \svpname{} to connect near-term interaction
dynamics with longer-term scene evolution and task intent. The same two scales
give the IDM fine temporal resolution over the imminent action chunk and
sufficient reach to determine task-level intent. Predicting every frame at
control rate over the full horizon,
however, would spend most of the token budget on the far horizon, where fine
resolution is unnecessary. We therefore adopt the multi-scale temporal design
of Act2Goal~\citep{zhou2025act2goal}. From the current frame $f_0$, the \svpname{}
predicts $N_d$ \emph{dense} frames uniformly strided across the
action-execution horizon $H$, and $N_s$ \emph{sparse} frames partitioning the
remainder of the clip up to its end frame $f_0 + T$,
\begin{equation}\label{eq:schedule}
\mathcal{F} \;=\; \{\,f_0\,\}
\;\cup\;
\underbrace{\Bigl\{\, f_0 + j\,\tfrac{H}{N_d} \,\Bigr\}_{j=1}^{N_d}}_{\text{dense: control}}
\;\cup\;
\underbrace{\Bigl\{\, f_0 + H + \Bigl\lfloor k\,\tfrac{T - H}{N_s} \Bigr\rceil \Bigr\}_{k=1}^{N_s}}_{\text{sparse: intent}}.
\end{equation}
The sparse partition is arithmetic and includes the clip end $f_0 + T$ at
$k=N_s$, with the reach $T$ set per clip. Default values of
$(N_d,N_s,H)$ are given in Appendix~\ref{app:recipes}.

\subsubsection{Toward Unified Understanding and Generation}
\label{sec:understanding_generation}

Multimodal modeling is increasingly moving toward systems that connect
understanding and generation instead of treating them as isolated
capabilities~\citep{xie2024showo,wu2024janus}. This connection is particularly
important for a visual planner: an instruction such as ``place it behind
the red bowl'' cannot be resolved from language alone, because its referents
and spatial relations depend on the current scene. A common video-generation
design, exemplified by Wan~\citep{wan2025}, represents language with a
text-only encoder and leaves this scene-specific grounding to the generative
backbone.

To provide scene-grounded conditioning, the \svpname{} uses a frozen vision--language model (VLM) as its
multimodal understanding component. The VLM jointly processes the current
head-view observation $o_{\mathrm{head}}$ and instruction $c$, producing hidden
states at each layer,
\begin{equation}\label{eq:vlmenc}
\bigl(h_0,h_1,\ldots,h_L\bigr)
= \mathrm{VLM}\bigl(o_{\mathrm{head}},c\bigr),
\qquad h_\ell \in \mathbb{R}^{S\times d}.
\end{equation}
We instantiate this component with Qwen3.5~\citep{qwen35blog}. Only the
text-span states $h_\ell[\mathcal{T}]$ are passed to the flow generator, but they
have been contextualized by the accompanying image. This keeps the
cross-attention sequence text-sized while grounding its content in the
observed scene.

Different VLM layers retain different mixtures of lexical, spatial, and visual
information. We therefore fuse their text-span states with an input-dependent
gate. Writing $g$ for the learned gating network and $\operatorname{Pool}$ for
masked token pooling,
\begin{equation}\label{eq:vlmgate}
\alpha_\ell
= \operatorname{softmax}_{\ell}
\!\left(g\!\left(\operatorname{Pool}
\bigl(h_\ell[\mathcal{T}]\bigr)\right)\right),
\qquad
c_{\mathrm{vlm}}
= \operatorname{LN}\!\left(\sum_{\ell=0}^{L}
\alpha_\ell h_\ell[\mathcal{T}]\right).
\end{equation}
Every DiT block cross-attends to $c_{\mathrm{vlm}}$, shared across generated
views. The two modules retain specialized backbones, but form a single
visual-planning pathway: the VLM resolves what the instruction refers to, and the
generative DiT predicts how that grounded scene should evolve. The IDM receives
neither $c$ nor $c_{\mathrm{vlm}}$ directly, so language affects action through
the generated future visual states and does not become a prerequisite for IDM pretraining.

\subsubsection{Conditional MeanFlow for One-Step Visual Planning}
\label{sec:conditional_meanflow}

The \svpname{} formulates temporally structured, conditional multi-view visual
prediction as a one-step MeanFlow problem~\citep{geng2025meanflow,geng2026improvedmeanflows}.
The resulting objective maps Gaussian noise to a complete future visual-latent sequence in one
differentiable forward pass. For recorded future visual latents $z$ and Gaussian noise
$\epsilon\sim\mathcal{N}(0,I)$, the linear trajectory is
$z_t=(1-t)z+t\epsilon$, with velocity $v_c=\epsilon-z$, and its average
velocity over an interval $[r,t]$ is
\begin{equation}\label{eq:meanvel}
u(z_t,r,t)
\mathrel{:=}
\frac{1}{t-r}\int_r^t v(z_\tau,\tau)\,\mathrm{d}\tau,
\end{equation}
which directly relates two points on the same trajectory through
$z_r=z_t-(t-r)u(z_t,r,t)$. With $(r,t)=(0,1)$, one forward evaluation of the learned
field maps noise to data. Differentiating Eq.~\eqref{eq:meanvel} with respect
to $t$ gives
\begin{equation}\label{eq:identity}
u(z_t,r,t)
= v(z_t,t)
- (t-r)\frac{\mathrm{d}}{\mathrm{d}t}u(z_t,r,t),
\end{equation}
where the total derivative is
\begin{equation}\label{eq:jvp}
\frac{\mathrm{d}}{\mathrm{d}t}u_\theta
= \partial_z u_\theta\cdot\frac{\mathrm{d}z_t}{\mathrm{d}t}
+ \partial_t u_\theta.
\end{equation}

We optimize this identity with a shared DiT trunk, a mean-velocity head
$u_\theta$, and an auxiliary instantaneous-velocity head $v_\theta$. The
auxiliary prediction supplies the state tangent in the total derivative, and a
single Jacobian--vector product (JVP) yields the compound prediction
\begin{equation}\label{eq:compound}
V_\theta
=u_\theta
+(t-r)\,\operatorname{sg}\!\left[
\frac{\mathrm{d}u_\theta}{\mathrm{d}t}\right],
\end{equation}
where $\operatorname{sg}[\cdot]$ stops gradients through the JVP correction.
The prediction $V_\theta$ is regressed against $v_c$, while $v_\theta$ receives
an auxiliary flow-matching loss.

To construct the conditional multi-view trajectory, we keep observed frames
fixed and evolve only future frames. Each training frame is encoded by
\coaename{} into a latent $z^{(i)}$ and assigned a binary mask
\begin{equation}\label{eq:mask}
m_i=
\begin{cases}
0, & \text{conditioning frame},\\[2pt]
1, & \text{future frame to predict}.
\end{cases}
\end{equation}
Suppressing the view axis, the conditional path is
\begin{equation}\label{eq:conditional_path}
z_t^{(i)}
=(1-m_i)z^{(i)}
+m_i\bigl((1-t)z^{(i)}+t\epsilon^{(i)}\bigr),
\qquad
v_c^{(i)}=m_i\bigl(\epsilon^{(i)}-z^{(i)}\bigr).
\end{equation}
Thus, conditioning coordinates remain clean and have zero velocity throughout
the path.

For the derivative to remain consistent with this partial trajectory, we mask
the predicted state tangent and the per-token interval endpoints,
\begin{equation}\label{eq:tangent}
\frac{\mathrm{d}z_t}{\mathrm{d}t}
=m\odot v_\theta(z_t,t),
\qquad
(r_i,t_i)=(m_i r,m_i t),
\end{equation}
so conditioning tokens contribute neither an advection tangent nor a time
derivative. The masked derivative is computed in one JVP computation. Both
heads are supervised only on prediction coordinates,
\begin{equation}\label{eq:loss}
\mathcal{L}_u
=\bigl\|m\odot(V_\theta-v_c)\bigr\|^2,
\qquad
\mathcal{L}_v
=\bigl\|m\odot(v_\theta-v_c)\bigr\|^2.
\end{equation}
Sampling distributions for $(r,t)$, dense/sparse weighting, and stabilization
details are given in Appendix~\ref{app:recipes}.

At inference, conditioning slots contain their clean latents and prediction slots
contain pure noise. A single forward pass at $(r,t)=(0,1)$ gives
\begin{equation}\label{eq:onestep}
\hat z^{(i)}
=z_1^{(i)}
-m_i\,u_\theta^{(i)}\bigl(z_1,r{=}0,t{=}1\bigr).
\end{equation}
Conditioning frames pass through unchanged, while predicted frames are fully
denoised in one forward pass. The resulting future visual latents are differentiable with
respect to all trainable flow-generator parameters, providing the explicit
world--action interface used in Section~\ref{sec:validity}.

\raggedbottom
\subsection{World-Action Alignment and \methodname{}}
\label{sec:validity}

The final part of pretraining first optimizes the IDM,
then connects the separately optimized \svpname{} and the IDM for co-training.
Joint optimization introduces conflicting video and action supervision,
including a video--action mismatch that we call \gapname{}
(Section~\ref{sec:validity-gap}).
We address it with Knowledge-Aligned Selective Optimization
(\methodname{}; Section~\ref{sec:kaso}),
which preserves action diversity in the resulting pretrained policy
and gives subsequent post-training methods such as reinforcement learning
broader coverage of the action space.

\subsubsection{Inverse Dynamics Model}
\label{sec:idm}

The IDM translates visual transitions into actions.
We denote the CoAE latents of the current multi-view observation by $o$,
the future visual latents by $z$,
and the current proprioceptive state by $s$.
The IDM attends to the future visual latents through cross-attention;
the future visual-state sequence spans the dense and sparse portions
of the temporal schedule in Section~\ref{sec:mstp}.
Conditioned on the future visual-state sequence,
the IDM jointly produces a dense control-rate action chunk
and sparse actions over the longer horizon.
At deployment, the controller executes only the dense action chunk.
The sparse actions are auxiliary training targets that encourage the IDM
to preserve longer-horizon intent.
With $a_t = t\,\varepsilon + (1-t)\,a$
denoting a noised action sequence at flow time $t$,
its velocity head $v_\phi$ is trained by
\begin{equation}\label{eq:idm}
\ell_{\mathrm{FM}}\bigl(a \,;\, o,\,s,\,z\bigr)
\;=\;
\mathbb{E}_{t,\varepsilon}
\Bigl\|
v_\phi\bigl(a_t,t;\,o,s,z\bigr)
- \bigl(\varepsilon-a\bigr)
\Bigr\|^2 .
\end{equation}
During separate IDM pretraining,
$a$ and the recorded future visual latents $z$ come from the same physical rollout,
so they are compatible by construction.
During co-training, $a$ remains the action stored in that rollout,
whereas the IDM receives generated future visual latents $\hat z$ from the \svpname.

\subsubsection{Knowledge Transfer with Conflicting Supervision}
\label{sec:derivation}

GE-Act~2.0 is pretrained in three steps.  \coaename{} is trained first and
then frozen, fixing the shared visual latent space.  In that space, the
\svpname{} is pretrained on instruction--video data and the IDM is pretrained
separately on paired recorded future--action data, so neither component
requires the other's annotations.  \methodname{} co-training then connects the
two pretrained checkpoints through generated futures while retaining their
original objectives.

Using these pretrained components, our one-step generator allows the
\svpname{} and the IDM to be co-trained directly through a
differentiable generator--IDM interface.
Let $G_\theta$ denote the one-step generation map in Eq.~\eqref{eq:onestep}.
The generated future and End-to-End Co-training (E2E) objective are
\begin{equation}\label{eq:e2e}
\hat z = G_\theta(o,c,\xi),
\qquad
\mathcal{L}_{\mathrm{E2E}}
= \ell_{\mathrm{FM}}(a;o,s,\hat z).
\end{equation}
The action gradient passes through the generated video to update both components.
However, optimizing only $\mathcal{L}_{\mathrm{E2E}}$ can overwrite capabilities
acquired during separate pretraining.
A similar effect has been reported in VLA training,
where optimizing only the action objective can degrade pretrained VLM capabilities,
motivating the continued use of VLM pretraining objectives during robot
training~\citep{driess2025knowledge}.
We therefore retain the \svpname{}'s video objective
$\mathcal{L}_{\mathrm{SVP}}=\mathcal{L}_u+\mathcal{L}_v$
and the IDM's recorded-video objective
$\mathcal{L}_{\mathrm{IDM}}=\ell_{\mathrm{FM}}(a;o,s,z)$ during co-training.
We refer to this variant as E2E with pretraining losses (E2E+PT).
Suppressing fixed loss coefficients,
\begin{equation}\label{eq:e2e_pt}
\mathcal{L}_{\mathrm{E2E+PT}}
= \mathcal{L}_{\mathrm{E2E}}
+ \mathcal{L}_{\mathrm{SVP}}
+ \mathcal{L}_{\mathrm{IDM}}.
\end{equation}
\begin{figure}[H]
\centering
\includegraphics[width=\linewidth]{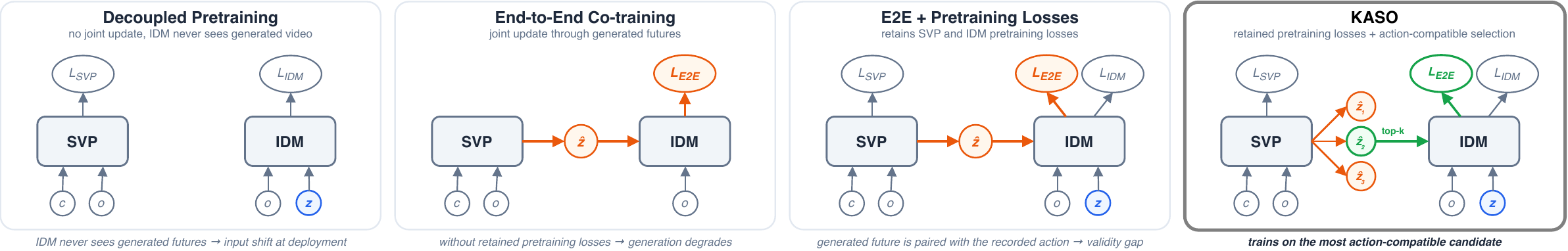}
\caption{Comparison of Decoupled Pretraining, End-to-End Co-training,
E2E + Pretraining Losses, and \methodname{}. Decoupled Pretraining is the
pretraining stage itself: the \svpname{} and the IDM are trained on their own
objectives and never exchange generated futures. The four strategies differ in
whether the IDM receives generated future visual states, whether the pretraining
losses for the \svpname{} and IDM remain active, and whether generated candidates are
filtered for action compatibility.}
\label{fig:training_patterns}
\end{figure}

The first three panels of Figure~\ref{fig:training_patterns}
compare Decoupled Pretraining, E2E, and E2E+PT.
E2E+PT retains both pretraining losses,
but it does not eliminate conflicting supervision during co-training.

\subsubsection{The Validity Gap}
\label{sec:validity-gap}

The remaining conflict arises within the generated co-training pair itself:
the IDM conditions on the generated $\hat z$,
while its target remains the recorded action $a$.
Although $a$ and the recorded $z$ come from the same rollout,
$\hat z$ is sampled separately
and need not depict a future consistent with $a$.

Manipulation is inherently multimodal:
given the same observation $o$ and instruction $c$,
the task can still be completed in several different ways.
The robot may approach from the left or the right,
grasp now or a moment later,
or choose among multiple objects that satisfy the instruction.
We refer to each behaviorally distinct realization
as an \emph{outcome mode}.
Under behavior cloning,
each demonstration provides action supervision only
for the outcome mode realized by its recorded episode,
whereas stochastic generation may sample any outcome mode
consistent with the same observation and instruction.
Consequently, $(a,z)$ share an outcome mode by construction,
while an independently generated $\hat z$ and the recorded $a$ need not.

Formally, the dataset provides a joint sample,
whereas the generated training pair combines two conditional marginals,
\begin{align}\label{eq:gap}
(z,a)
&\sim p_{\mathrm{data}}(z,a\mid o,s,c),
&&\text{recorded in the same rollout},
\nonumber\\[-1pt]
\hat z &\sim p_\theta(z\mid o,c),
\qquad
a \sim p_{\mathrm{data}}(a\mid o,s,c),
\qquad
\hat z \perp a \mid (o,s,c),
&&\text{paired during co-training}.
\end{align}
A generated pair is action-compatible when $\hat z$ and $a$
instantiate the same outcome mode;
otherwise the IDM is asked to predict one mode's action
while conditioning on another mode's future visual states.
We call this video--action mismatch \textbf{\gapname{}}.
The conditional independence in Eq.~\eqref{eq:gap}
distinguishes the gap from generation error:
even if $p_\theta(z\mid o,c)$ matches the true future marginal,
better generation makes each sampled $\hat z$ more plausible
but does not couple its outcome mode to the recorded action.

\paragraph{A Controlled Toy Two-Stage System.}
To isolate the consequences of \gapname{},
we construct a toy two-stage system
(details in Appendix~\ref{app:toy})
and train it with each strategy in Figure~\ref{fig:training_patterns},
starting from identical initializations.
Future visual states are drawn from an equal-weight mixture
of four Gaussian outcome modes centered at
$(\pm1.5,\pm1.8,0)$, with every recorded sample lying exactly on the
$z_3=0$ plane.
Actions follow the deterministic mapping
$a=f(z)=z_1z_2/(1.5\times1.8)$,
which maps the four video modes to two action modes near $a=\pm1$.
Consequently, an IDM trained only on recorded pairs is unconstrained away from
the ground-truth plane when a generator produces residual $z_3$ deviations.
To expose this off-manifold failure mode, every arm uses the same IDM input
normalization $(1.5,1.8,0.2)$; the third-coordinate scaling leaves all recorded
inputs unchanged but magnifies generated off-plane deviations equally across
methods.  Full controls and training details are given in
Appendix~\ref{app:toy}.
We use a flow-matching generator as the first stage of this system,
corresponding to the \svpname{} in the full WAM,
and an action regressor as the second stage,
serving as a simplified IDM.

\begin{figure}[H]
\centering
\includegraphics[width=\linewidth]{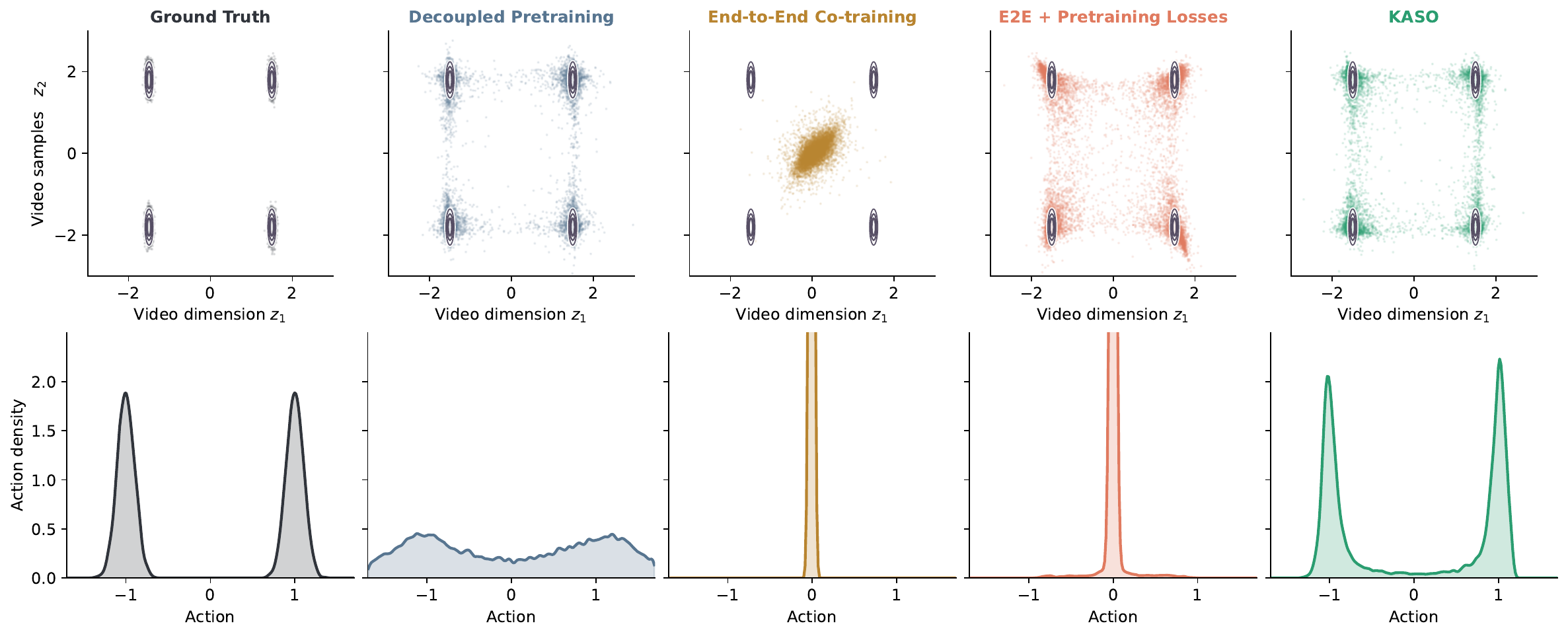}
\caption{Video samples projected onto the $z_1$--$z_2$ plane
(top, with ground-truth contours overlaid)
and induced action densities (bottom) learned by each strategy
on the toy system.  The displayed run uses seed~0 and single-sample evaluation
for every learned strategy; seed-to-seed variation is discussed in
Appendix~\ref{app:toy}.
Ground truth contains four video modes and two action modes;
in the displayed run, Decoupled Pretraining preserves the four video modes but yields
a diffuse action density;
End-to-End Co-training collapses both the video and action distributions;
with the deterministic IDM, E2E + Pretraining Losses preserves the video modes
but collapses the action density to the mean of the two action modes;
and \methodname{} recovers the four video modes
and the bimodal action structure.}
\label{fig:toy_cascade}
\end{figure}

Figure~\ref{fig:toy_cascade} shows the learned video distribution
and its induced action density for seed~0.
Ground truth contains four video modes
and two action modes near $\pm1$.
For this displayed run, Decoupled Pretraining covers all four video modes,
but its action density is diffuse:
its IDM is trained only on recorded states
and is therefore unconstrained on generated states outside that support.
Because generated futures and recorded actions are paired independently
in Eq.~\eqref{eq:gap}, both E2E and E2E+PT associate
each generated outcome mode with equally weighted action targets near $\pm1$.
Under squared-error regression, the IDM regressor in both methods
minimizes this conflicting supervision by predicting their mean at zero.
Without the retained pretraining losses, E2E also loses
the four video modes retained by E2E+PT.
A generative IDM may preserve both action modes,
but the same pairing still removes their dependence
on the outcome mode shown by the generated video.
Our proposed method, \methodname{}, described next,
couples the \svpname{} and the IDM through action-compatible pairs
and recovers both the multimodal video distribution
and the bimodal action distribution,
so subsequent post-training begins with a more diverse action distribution,
which is particularly important for methods such as reinforcement learning.
The toy \methodname{} arm selects one of eight candidates during training only;
all arms use one generated sample per evaluation input.  Moreover, at seed~0
the Decoupled arm has the lowest IDM error on recorded pairs and the lowest
video sliced-Wasserstein distance among the learned arms, localizing its failure
to the generated-video action interface rather than either pretrained module.

\FloatBarrier

\subsubsection{\methodname{}: Knowledge-Aligned Selective Optimization}
\label{sec:kaso}

\begin{figure}[!t]
\centering
\includegraphics[width=\linewidth]{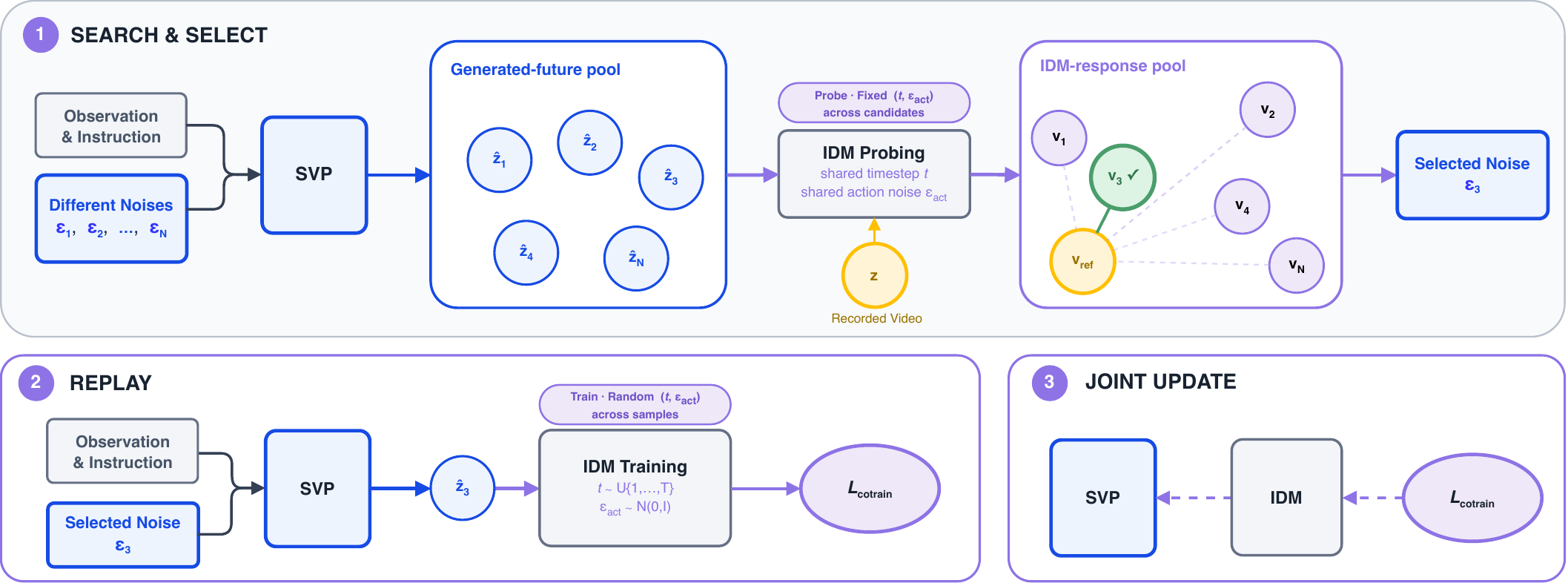}
\caption{The \methodname{} pipeline.  Candidate visual predictions are sampled
from the \svpname{} and scored without gradients; selected noises are then
replayed exactly so that the joint action loss updates both the IDM and the
\svpname{}.}
\label{fig:kaso_algorithm}
\end{figure}

Therefore, we propose \methodname{},
an online top-$k$ procedure for selecting
action-compatible future visual-state sequences during co-training,
as illustrated in Figure~\ref{fig:training_patterns}.
At a high level, for each context, the \svpname samples $N$ candidate future videos.
The active IDM translates each candidate into an action response
and compares it with its response to the recorded video.
\methodname{} selects the top-$k$ candidates with the smallest discrepancy;
only those candidates are replayed with gradients
and receive the generated-video action loss
(Figure~\ref{fig:kaso_algorithm} and Algorithm~\ref{alg:kaso}).
The comparison is deliberately made in action space rather than pixel space:
two visually different future visual-state sequences may support the same action,
for example when they differ only in task-irrelevant background motion,
whereas a visually convincing prediction from another mode does not.
In the toy system of Figure~\ref{fig:toy_cascade},
compatibility-based selection suppresses mismatched cross-mode pairs,
allowing \methodname{} to retain all four video modes
and recover the two action modes near $\pm1$.

Relative to E2E+PT in Eq.~\eqref{eq:e2e_pt},
\methodname{} changes only the generated-video action branch
by applying it to the selected top-$k$ candidates.
Comparing E2E+PT with \methodname{} therefore isolates selection,
while comparing E2E with E2E+PT isolates the two retained pretraining losses.

To score the candidates, \methodname{} probes the IDM
at a fixed high-noise point $t_p$ on its action-flow path.
It draws one probe noise $\varepsilon$ and constructs
$\tilde a=t_p\varepsilon+(1-t_p)a$;
the same $\tilde a$ is then evaluated under the recorded future visual states
and under every generated candidate.
Sharing the action input and the active IDM parameters
leaves the future visual conditioning as the only changing input.
For the set $\mathcal{D}$ of dense action tokens,
the candidate energy is
\begin{equation}\label{eq:energy}
E_n
= \frac{1}{|\mathcal{D}|}
\sum_{i\in\mathcal{D}}
\Bigl\|
v_{\phi,i}\bigl(\tilde a,t_p;\,o,s,\hat z_n\bigr)
-
v_{\phi,i}\bigl(\tilde a,t_p;\,o,s,z\bigr)
\Bigr\|_2^2 .
\end{equation}
We average only over dense tokens,
which directly cover the action-execution horizon
depicted by the near-horizon frames.
Probing at high noise is what makes this energy informative.
At low noise, $\tilde a\approx a$ and the flow target $\varepsilon-a$
is largely predictable from the action input alone,
so the velocity depends only weakly on the future visual conditioning
and all candidates receive similar energies.
At high noise, $\tilde a$ carries almost no information about $a$,
so the velocity must be inferred from $(o,s,\hat z_n)$,
and Eq.~\eqref{eq:energy} reduces to the disagreement between
the action the IDM reads from the candidate future
and the action it reads from the recorded future.
A low energy therefore means that the current IDM
cannot behaviorally distinguish the candidate from the recorded future,
which is the action-compatibility notion introduced above.

Let $\mathcal{S}_k$ contain the indices of the $k$ candidates
with the lowest energies.  The selected E2E loss is
\begin{equation}\label{eq:sel}
\mathcal{L}_{\mathrm{E2E}}^{\mathrm{sel}}
= \frac{1}{k}\sum_{n\in\mathcal{S}_k}
\ell_{\mathrm{FM}}(a;o,s,\hat z_n).
\end{equation}
The full \methodname{} objective retains the same two pretraining losses:
\begin{equation}\label{eq:kaso_objective}
\mathcal{L}_{\mathrm{KASO}}
= \mathcal{L}_{\mathrm{E2E}}^{\mathrm{sel}}
+ \mathcal{L}_{\mathrm{SVP}}
+ \mathcal{L}_{\mathrm{IDM}}.
\end{equation}
We use $k=1$ in our implementation.

Candidate generation and scoring are performed without gradients,
while the generation noise for every candidate is retained.
After selection, the \svpname replays the selected noises with gradients enabled;
because generation is a deterministic one-step map of each noise,
every replay reconstructs exactly the visual prediction that was scored.
The IDM then applies its standard flow-matching objective
with a freshly sampled action noise and flow time,
and the gradient passes through both the IDM and the \svpname
via Eq.~\eqref{eq:onestep}.
The update also includes $\mathcal{L}_{\mathrm{SVP}}$
on the recorded video
and $\mathcal{L}_{\mathrm{IDM}}$ on recorded action--video pairs.
Thus, \methodname{} adapts the IDM to generated inputs,
shapes the \svpname{}'s flow generator with the action objective,
and continues both pretrained objectives in the same stage.

Selection is recomputed online at every optimizer step
using the current \svpname{} and active IDM,
so the selected examples track both models as they change.
It also relies on a stochastic video generator:
without diversity across noise draws,
the candidate pool would contain no alternative mode to select.
This online resampling distinguishes \methodname{} from offline data-curation
pipelines that generate and filter a fixed synthetic dataset before policy
training~\citep{kim2026robocurate}.
Group size $N$, probe level $t_p$,
loss coefficients, and the recorded/generated mixing ratio
are given in Appendix~\ref{app:recipes}.

\begin{algorithm}[H]
\caption{One co-training step with \methodname{}.
Setting $k=N$ recovers E2E+PT;
removing in addition $\mathcal{L}_{\mathrm{SVP}}$
and $\mathcal{L}_{\mathrm{IDM}}$ recovers E2E.}
\label{alg:kaso}
\begin{algorithmic}[1]
\Require \svpname{} flow-generator parameters $\theta$, IDM parameters $\phi$
\Require Current observation latents $o$, proprioceptive state $s$,
  instruction $c$, recorded future visual latents $z$, recorded action $a$
\Require Group size $N$, selected count $k$, probe noise level $t_p$
\Statex
\Statex \textit{// Pretraining losses on recorded data}
\State Sample video flow times and noise;
  compute $\mathcal{L}_{\mathrm{SVP}}=\mathcal{L}_u+\mathcal{L}_v$
  \hfill \textit{Eq.~\eqref{eq:loss}}
\State $\mathcal{L}_{\mathrm{IDM}} \gets
  \ell_{\mathrm{FM}}(a;\,o,s,z)$
  \hfill \textit{Eq.~\eqref{eq:idm}}
\Statex
\Statex \textit{// Candidate generation without gradients}
\For{$n=1,\ldots,N$}
  \State Draw and retain generation noise $\xi_n$
  \State $\hat z_n \gets$ one-step generation from $(o,c,\xi_n)$
    \hfill \textit{Eq.~\eqref{eq:onestep}}
\EndFor
\Statex
\Statex \textit{// Compatibility probe with the active IDM, no gradients}
\State Draw shared probe noise $\varepsilon$;
  $\tilde a \gets t_p\varepsilon+(1-t_p)a$
\State $v_{\mathrm{ref}} \gets
  v_\phi(\tilde a,t_p;\,o,s,z)$
\For{$n=1,\ldots,N$}
\State $E_n \gets
    \|v_\phi(\tilde a,t_p;\,o,s,\hat z_n)
    -v_{\mathrm{ref}}\|_{\mathrm{dense}}^2$
    \hfill \textit{Eq.~\eqref{eq:energy}}
\EndFor
\Statex
\Statex \textit{// Top-$k$ selection and differentiable replay}
\State $\mathcal{S}_k \gets$ indices of the $k$ smallest energies in $\{E_n\}_{n=1}^N$
\For{$n\in\mathcal{S}_k$}
  \State $\hat z_n \gets$ replay $\xi_n$ with gradients
\EndFor
\Statex
\Statex \textit{// Joint update}
\State $\mathcal{L}_{\mathrm{E2E}}^{\mathrm{sel}} \gets
  \frac{1}{k}\sum_{n\in\mathcal{S}_k}
  \ell_{\mathrm{FM}}(a;\,o,s,\hat z_n)$
  \hfill \textit{Eq.~\eqref{eq:sel}}
\State $\mathcal{L}_{\mathrm{KASO}} \gets
  \lambda_{\mathrm{E2E}}\mathcal{L}_{\mathrm{E2E}}^{\mathrm{sel}}
  +\lambda_{\mathrm{SVP}}\mathcal{L}_{\mathrm{SVP}}
  +\lambda_{\mathrm{IDM}}\mathcal{L}_{\mathrm{IDM}}$
  \hfill \textit{Eq.~\eqref{eq:kaso_objective}}
\State $\theta,\phi \gets \theta,\phi
  -\alpha\nabla_{\theta,\phi}\mathcal{L}_{\mathrm{KASO}}$
\end{algorithmic}
\end{algorithm}

\subsubsection{Controlled Real-Robot Ablation}
\label{sec:evidence}

To isolate the effect of the alignment objective, we compare E2E, E2E+PT,
and \methodname{} under a deliberately reduced, matched alignment-stage and
evaluation setup.  Every variant initializes from the same full-scale
components: the \svpname{}, pretrained on \svpPretrainHours{} hours of
instruction--video data, and the IDM, pretrained on \idmPretrainHours{} hours of
action-labeled trajectories.  Only the connection/alignment stage is run on a
300-hour co-training mixture, including 30 hours from G2-90D, the embodiment used
for evaluation.  Thus, this is a 300-hour alignment stage atop full-scale
pretrained components, not a model trained on 300 hours in total.  The exact
evaluation environment and physical object instances are unseen during
training, and no evaluation-specific SFT is
performed.  We evaluate two real-robot picking protocols.  In the single-object
protocol, the same small block is placed at different table positions for 25
trials.  In the four-object protocol, a sponge, red apple, computer mouse, and
remote control are presented simultaneously, and the robot is instructed to
pick each designated target in 10 trials.

We report two complementary metrics.  In the four-object protocol, the follow
score counts trials in which the robot contacts the instructed target and no
distractor object, regardless of whether the subsequent grasp and lift
succeed.  Pick success additionally requires completing the pickup.  Because
the single-object protocol contains no distractor, follow score is not defined
for that protocol.

\begin{table}[H]
\captionsetup{skip=8pt}
\caption{Controlled real-robot ablation of retaining pretraining losses and
action-compatible selection.  Follow score requires contacting the instructed
target and no distractor; pick success additionally requires completing the
pickup.  Four-object macro averages assign equal weight to the four targets.
All variants initialize from the same full-scale pretrained components
(\svpname{}, \svpPretrainHours{} hours; IDM, \idmPretrainHours{} hours) and use
the same 300-hour connection/alignment mixture, including 30 hours from G2-90D,
without evaluation-specific SFT.  Entries are percentages.}
\label{tab:kaso_ablation}
\centering
\tablebodyfont
\setlength{\tabcolsep}{4.8pt}
\begin{tabular}{lcccccc}
\toprule
\multicolumn{7}{l}{\textit{(a) Follow score: four-object scene} (10 trials per target)} \\
\addlinespace[1pt]
\multicolumn{2}{l}{Method} & Sponge & Red apple & Mouse & Remote & Macro avg. \\
\midrule
\multicolumn{2}{l}{E2E} & 90 & 100 & 90 & 70 & 87.5 \\
\multicolumn{2}{l}{E2E+PT} & 100 & 100 & 90 & 90 & 95 \\
\rowcolor{RedOrange!8}
\multicolumn{2}{l}{\textbf{\methodname{}}} & 100 & 100 & 90 & 90 & 95 \\
\addlinespace[5pt]
\multicolumn{7}{l}{\textit{(b) Pick success}} \\
\addlinespace[1pt]
& \multicolumn{1}{c}{Single-object scene (25 trials)}
& \multicolumn{5}{c}{Four-object scene (10 trials per target)} \\
\cmidrule(lr){2-2}\cmidrule(lr){3-7}
Method & Small block & Sponge & Red apple & Mouse & Remote & Macro avg. \\
\midrule
E2E & 12 & 30 & 60 & 20 & 0 & 27.5 \\
E2E+PT & 12 & 40 & 40 & 10 & 0 & 22.5 \\
\rowcolor{RedOrange!8}
\textbf{\methodname{}} & 40 & 60 & 60 & 20 & 10 & 37.5 \\
\bottomrule
\end{tabular}
\end{table}

Table~\ref{tab:kaso_ablation} separates target following from completed pickup.
In the four-object protocol, E2E+PT and \methodname{} achieve the same 95\%
follow score, but \methodname{} converts more correct target contacts into
successful picks: its macro-average pick success is 37.5\%, compared with
22.5\% for E2E+PT ($+15$ percentage points), and it improves pick success for
every tested target.  In the single-object protocol, \methodname{} raises pick
success from 12\% to 40\% ($+28$ points).  Relative to E2E, \methodname{}
improves the four-object follow score from 87.5\% to 95\%, the four-object
macro-average pick success from 27.5\% to 37.5\%, and the single-object pick
success from 12\% to 40\%.

Table~\ref{tab:kaso_ablation} provides a controlled demonstration of
\methodname{} in a simple, matched setting.
By holding the full-scale pretrained initialization, 300-hour alignment
mixture, embodiment data, and evaluation protocol fixed, the comparison
isolates the effect of the alignment objective.  Accordingly, this table
compares 300-hour alignment stages atop full-scale pretrained components rather
than models trained from scratch on 300 hours.
This experiment is separate from the full-scale capability evaluation in
Section~\ref{sec:scaling}: the full model is trained with substantially more
data and exhibits stronger capability, so the values in
Table~\ref{tab:kaso_ablation} should not be interpreted as its final
performance.

\section{Training Data and Mixture Composition}
\label{sec:data}

\paraid{Data-P1}\sentid{1}GE-Act~2.0 uses three stage-specific data mixtures.
\sentid{2}\svpname{} pretraining uses
instruction--video data and can incorporate
manipulation recordings without robot action labels, including egocentric
video.  \sentid{3}IDM pretraining uses action-labeled robot trajectories,
including demonstrations, failure trajectories, and deployment data.
\sentid{4}Co-training with \methodname{} connects the two components through
instruction--video--action data while retaining their separately pretrained
objectives (Section~\ref{sec:validity}).

\begin{figure}[H]
  \centering
  \includegraphics[width=\linewidth]{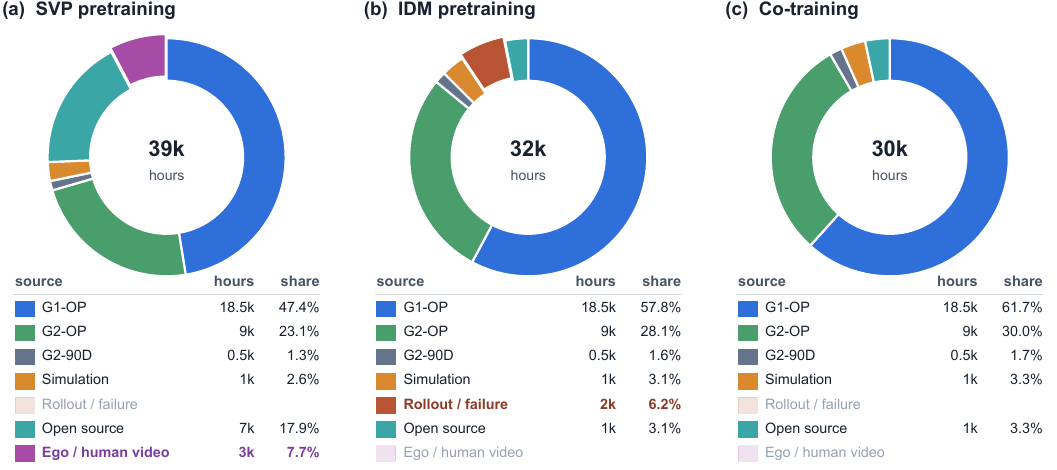}
  \caption{Stage-specific training-data mixtures.  The panels show the
  \svpPretrainHours-hour \svpname{} pretraining mixture, the
  \idmPretrainHours-hour IDM pretraining mixture, and the 30{,}000-hour
  co-training mixture.  Slice angle denotes the
  share of data hours; colors and source order are held fixed across panels.
  Violet and coral highlights emphasize the action-free egocentric/human video
  used for \svpname{} pretraining and the rollout/failure trajectories used for
  IDM pretraining, respectively.  The keys report source hours and percentages.}
  \label{fig:data_mixtures}
\end{figure}

\paraid{Data-P2}\sentid{1}The data span G1-OP, G2-OP, G2-90D, simulation, open-source
robot datasets, egocentric and human manipulation video, and rollout and
failure data.  \sentid{2}Their proportions differ across
\svpname{} pretraining, IDM pretraining, and co-training, as summarized in
Figure~\ref{fig:data_mixtures}.  \sentid{3}G1-OP contributes more than 50\% of the
co-training mixture, whereas G2-90D contributes less than 2\%.
\sentid{4}This imbalance motivates reporting atomic-skill results separately for
the two embodiments in Section~\ref{sec:scaling_skills}.
\sentid{5}The three stages contain \svpPretrainHours{}, \idmPretrainHours{}, and
30{,}000 hours, respectively.

\section{Evaluating Zero-Shot OOD Manipulation at Scale}
\label{sec:scaling}

\paraid{5-P1}\sentid{1}Our central evaluation resolves manipulation capability
into 100 atomic tasks, revealing which individual skills emerge,
improve, or remain difficult as training scales.  \sentid{2}We track this broad
repertoire across data scale on both G1-OP and G2-90D, with G1-OP
as the dominant embodiment in the co-training mixture and G2-90D as a data-scarce
embodiment contributing less than 2\% of it.  \sentid{3}We then
complement the atomic-skill evaluation with two controlled complex-scene
studies: a quantitative decomposition of grounding across object identity,
color, size, shape, position, and order, and qualitative stress tests in which
the current instruction conflicts with an ongoing behavior or a conventional
scene relation.  \sentid{4}The section concludes with three standardized
simulation benchmarks that test OOD conditions after each benchmark's
prescribed in-distribution adaptation; we interpret these results separately
from the direct real-robot zero-shot evaluation.  \sentid{5}Because
GE-Act~2.0 is designed as a low-level
manipulation policy rather than a high-level planner, these experiments focus
on executable visuomotor competence and fine-grained grounding rather than
high-level planning or open-ended reasoning.

\subsection{Zero-Shot OOD Evaluation without Per-Task SFT}
\label{sec:pretrained_eval}

\paraid{5.1-P1}\sentid{1}Existing world-action and
vision--language--action evaluations often include a task-specific adaptation
stage
\citep{zhang2026lingbot,zhang2026lingbotva2,gigaworldpolicy2026,zhou2026tau0wm,wu2024gr1,cheang2024gr2,kim2024openvla,black2024pi0}.  \sentid{2}Such evaluations measure
the resulting specialist policy and therefore answer a different question.
\sentid{3}In contrast, all our main real-robot experiments evaluate pretrained
checkpoints directly, without per-task SFT or any subsequent adaptation.
\sentid{4}This protocol isolates capabilities already present in the
pretrained policy rather than capabilities introduced through demonstrations
of the evaluation task.

\paraid{5.1-P2}\sentid{1}We use \emph{zero-shot} to mean that a checkpoint
receives no evaluation-task fine-tuning, demonstrations, or per-task
checkpoint selection.  \sentid{2}The evaluation is \emph{OOD}: the test scenes,
backgrounds, lighting configurations, and object instances are excluded from
pretraining and co-training.  \sentid{3}At each evaluation point, one checkpoint
is deployed across the complete suite with fixed deployment settings; full
protocol and checkpoint details are provided in
Appendix~\ref{app:recipes}.

\begin{figure}[p]
  \centering
  \vspace*{-11pt}
  \includegraphics[width=\linewidth,trim=32bp 15bp 32bp 20bp,clip]{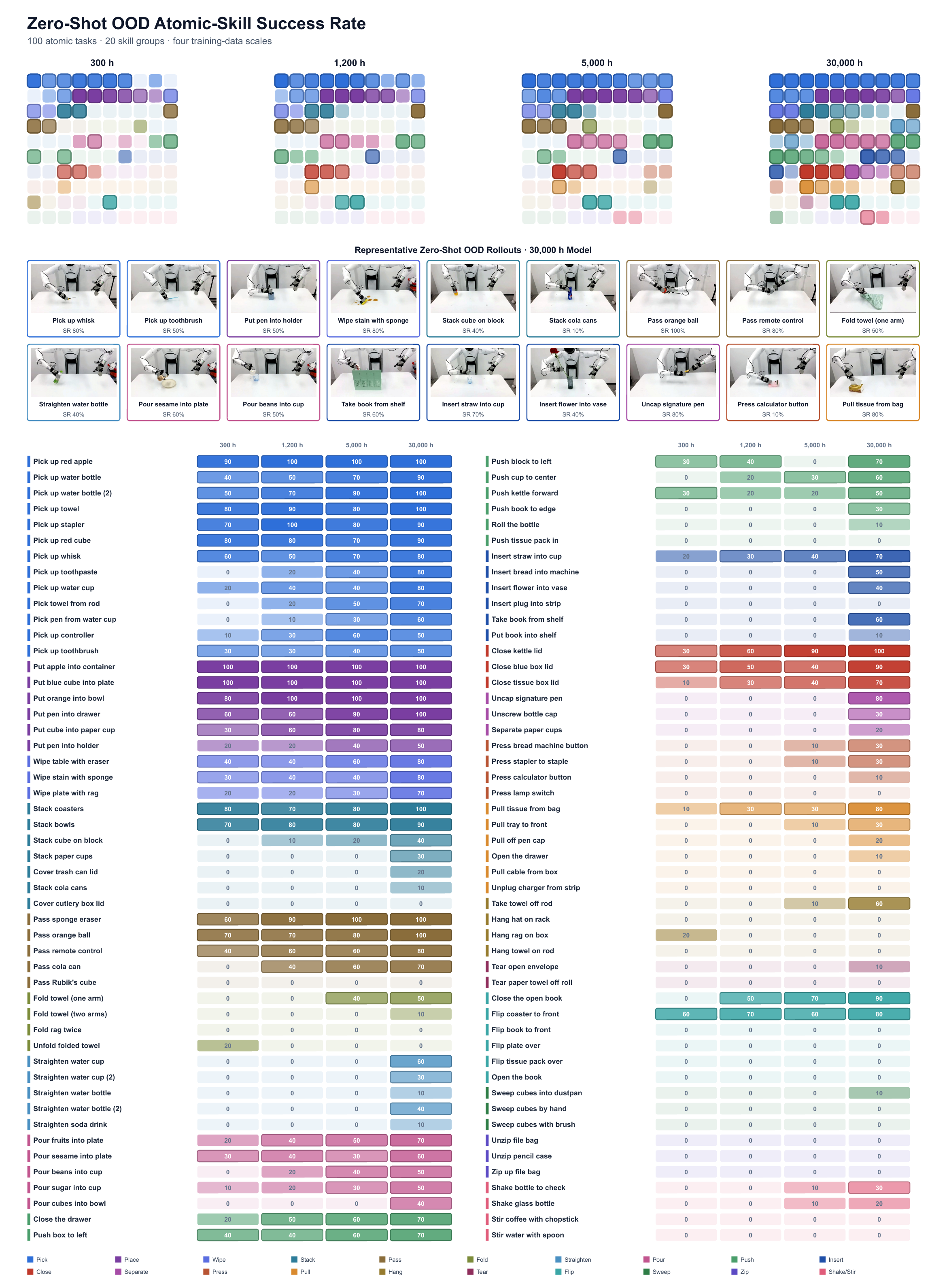}
  \captionsetup{font=scriptsize,skip=0pt}
  \caption{G1-OP zero-shot OOD scaling across 100 atomic tasks and four data
  volumes.  Top: the 100 tasks as one mosaic per training-data scale, with
  tile intensity denoting per-task SR.  Middle: representative zero-shot OOD
  rollouts of the 30{,}000-hour model.  Bottom: per-task SR at every scale,
  grouped by skill; colors follow the skill-group legend.}
  \vspace*{-10pt}
  \label{fig:data_scaling_wall}
\end{figure}

\subsection{Scaling Manipulation Skills}
\label{sec:scaling_skills}

\paraid{5.2-P1}\sentid{1}Atomic skills provide the most direct measure of an
executable low-level manipulation policy.  \sentid{2}They minimize confounding
from high-level planning and long-horizon error accumulation, allowing the
evaluation to focus on whether the policy can perceive a local manipulation
problem and execute the corresponding action.

\paraid{5.2-P2}\sentid{1}Our real-robot suite contains 100 tasks across 20 skill
groups, spanning pick, place, push, stack, tool use, articulated-object
manipulation, insertion, deformable-object manipulation, and fine motor
control.
\sentid{2}This breadth is intended to measure a repertoire rather than a small
set of canonical or headline tasks.  \sentid{3}Every task has a pre-specified
terminal success criterion, and the same suite is evaluated on G1-OP and
G2-90D.  \sentid{4}Each task is evaluated over 10 trials for every combination
of embodiment and training-data scale under a standardized reset procedure.
\sentid{5}We report empirical success rate (SR) at the task, skill-group, and
suite levels; the complete taxonomy, reset protocol, and raw results are
provided in Appendix~\ref{app:full_results}.

\subsubsection{Scaling with Training Data Volume}
\label{sec:data_scaling}

\paraid{5.2.1-P1}\sentid{1}A central question for large-scale robot pretraining
is whether expanding the training corpus produces a broader and more reliable
repertoire of zero-shot OOD manipulation skills.  \sentid{2}We train separate
models with nested co-training pools containing 300, 1{,}200, 5{,}000, and
30{,}000 hours.  \sentid{3}The smaller pools are contained in the larger ones
and preserve the overall mixture composition as closely as possible.
\sentid{4}All models start from the same pretrained components, are co-trained with
\methodname{} (Section~\ref{sec:kaso}), and use the same training recipe; each is trained until completing one epoch or reaching
the maximum duration permitted by our compute budget, whichever comes first.
\sentid{5}This is therefore a practical end-to-end scaling comparison rather
than a matched-compute isolation of training-data volume.

\begin{figure}[!t]
  \centering
  \includegraphics[width=\linewidth,trim=8bp 6bp 6bp 5bp,clip]{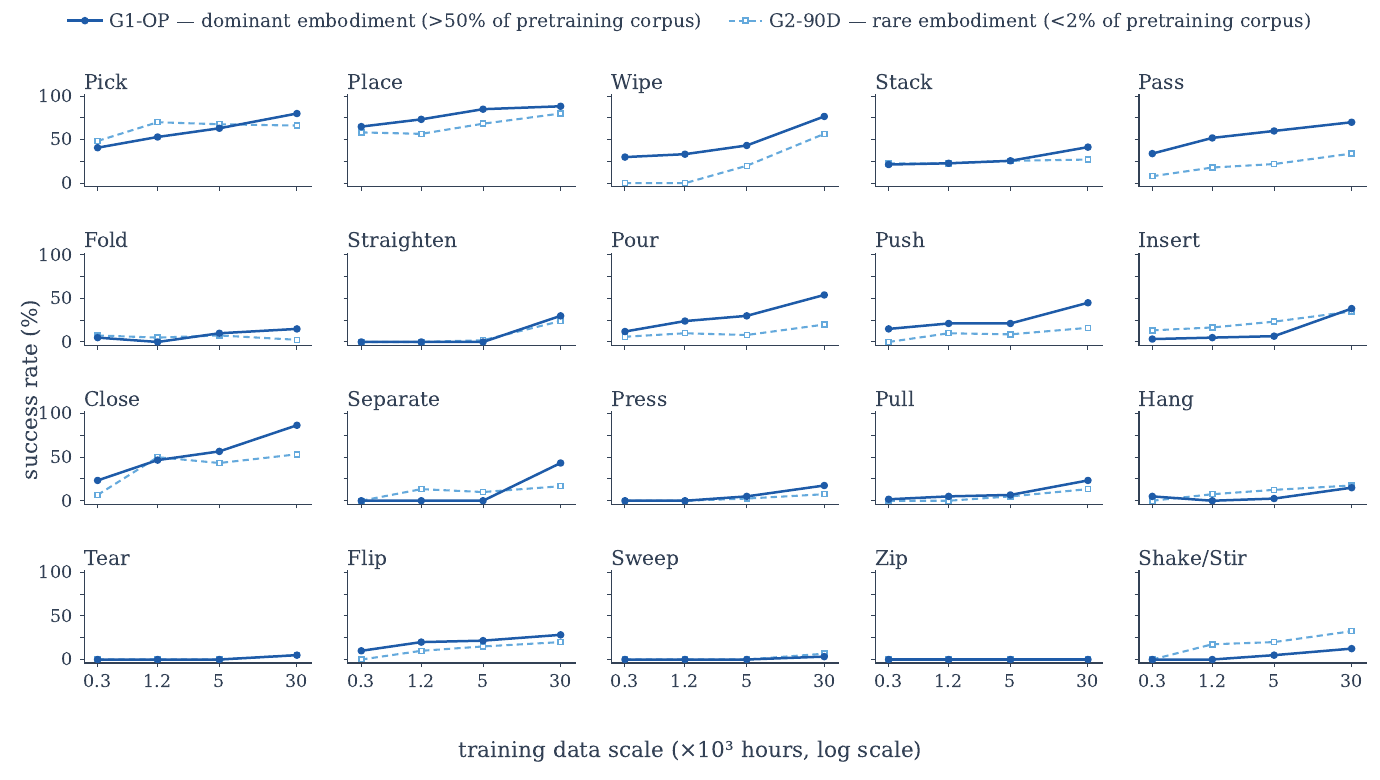}
  \caption{Skill-group zero-shot OOD scaling trajectories for G1-OP (more than
  50\% of the co-training mixture) and G2-90D (less than 2\%).  Each panel
  reports one group's mean SR across the 300-, 1{,}200-, 5{,}000-, and
  30{,}000-hour models.}
  \label{fig:data_scaling_families}
\end{figure}

\paraid{5.2.1-P2}\sentid{1}We report the results at three levels: task,
skill group, and suite.  \sentid{2}Figure~\ref{fig:data_scaling_wall} shows
every task as a wall for G1-OP, grouped into the 20 skill groups, with columns
denoting the four training-data scales.  \sentid{3}Figure~\ref{fig:data_scaling_families}
summarizes the same evaluation as paired G1-OP and G2-90D trajectories for all
20 skill groups.  \sentid{4}Suite-level SR for both embodiments is reported
below.

\begin{figure}[t]
  \centering
  \includegraphics[width=\linewidth]{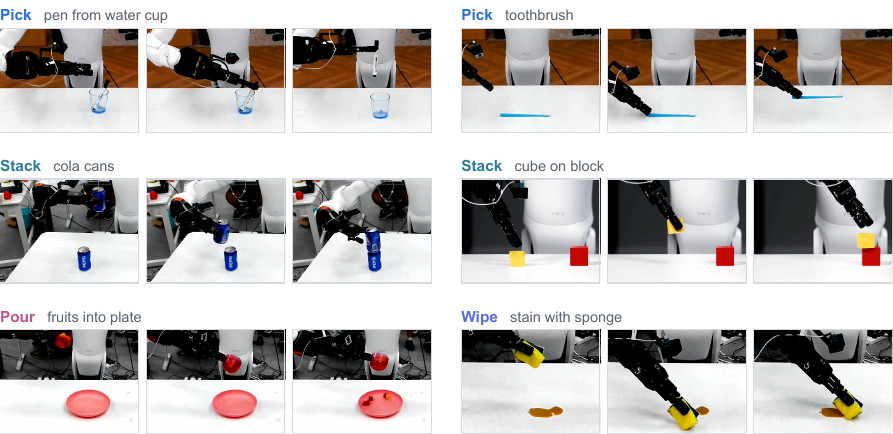}
  \caption{Representative successful zero-shot OOD rollouts on G2-90D across
  skill groups, each shown as three frames spanning the start, mid-execution,
  and outcome of the trial.  Skill labels are colored by skill group, matching
  the palette of Figure~\ref{fig:data_scaling_wall}.}
  \label{fig:g2_90d_rollout}
\end{figure}

\paraid{5.2.1-P3}\sentid{1}G1-OP suite-level SR increases from 17.1\% with 300
hours to 22.6\% with 1{,}200 hours, 27.3\% with 5{,}000 hours, and 44.1\% with
30{,}000 hours.  \sentid{2}G2-90D follows the same upward trend, increasing
from 13.4\% to 21.0\%, 23.5\%, and 31.1\% across the four data scales.
\sentid{3}These trajectories yield total gains of 27.0 and 17.7 percentage
points for G1-OP and G2-90D, respectively.  \sentid{4}Both curves continue to
rise substantially from 5{,}000 to 30{,}000 hours, providing no evidence of
saturation within the measured data range.

\paraid{5.2.1-P4}\sentid{1}The improvement is distributed broadly across the
repertoire: 19 of 20 G1-OP skill groups and 18 of 20 G2-90D groups perform
better at 30{,}000 hours than at 300 hours; the exceptions are Zip, with no
success in any of its 240 trials across both embodiments and all four scales,
and Fold on G2-90D, whose apparent decline spans three successful trials at
300 hours and one at 30{,}000.  \sentid{2}Zip is both mechanically demanding
and nearly absent from the training data: each of its three tasks requires a
precise pinch on a small slider and a steady pull along its track while the
other arm stabilizes the item, and the group receives only 16.4 hours across
the whole corpus, the second-lowest coverage of any skill group.
\sentid{3}Section~\ref{sec:skill_data_distribution} takes up this relationship
between coverage and capability across the full taxonomy.  \sentid{4}At the
atomic-task level, the number of tasks with nonzero empirical SR rises from 39
to 76 for G1-OP and from 24 to 72 for G2-90D.  \sentid{5}Across the four
scales, 87 of 100 G1-OP task trajectories and 78 of 100 G2-90D trajectories
are non-decreasing.  \sentid{6}Excluding the mechanically trivial Close group,
the largest G1-OP gains include Wipe ($+46.7$ points), Separate ($+43.3$),
Pour ($+42.0$), and Pick ($+39.2$).
\sentid{7}Scaling therefore improves most of the measured repertoire, although
the trajectories are not uniformly monotonic, particularly for G2-90D.

\paraid{5.2.1-P5}\sentid{1}\textbf{Scaling on the data-scarce embodiment.}
The 17.7-point gain on G2-90D is especially notable because it contributes less
than 2\% of the co-training mixture.  \sentid{2}Within the skill-attributed
G2-90D training data at the largest scale, 10 of the 20 skill groups contain
fewer than five hours, including six with fewer than two hours and four with
fewer than one hour.  \sentid{3}Nevertheless, five of the six groups with fewer
than two hours improve from zero SR with the 300-hour model to nonzero SR with
the 30{,}000-hour model.  \sentid{4}Flip and Separate reach 20.0\% and 16.7\%
SR with only 1.2 and 1.25 hours of G2-90D-specific data, respectively, while Pass
reaches 34.0\% with 3.9 hours.  \sentid{5}One hardware-sensitive exception is
Shake/Stir, which scores higher on G2-90D than on G1-OP.  \sentid{6}In rollout
inspection, G1-OP more often drops cups during shaking when its gripper fails to
maintain a stable hold on large containers, whereas G2-90D's wider gripper
provides a more secure grasp; this failure pattern suggests that the absolute
embodiment gap is partly attributable to end-effector geometry.  \sentid{7}This
hardware effect does not explain the within-G2-90D scaling trend, which compares
models on the same embodiment across data scales.  \sentid{8}Together with the
aggregate trend in Figure~\ref{fig:data_scaling_families}, these results suggest
that a sparsely represented embodiment can acquire substantially broader
capability as the shared multi-embodiment training corpus scales.
\sentid{9}Figure~\ref{fig:g2_90d_rollout} shows representative successful
G2-90D rollouts across skill groups.

\subsubsection{Scaling with Skill-Specific Data Volume}
\label{sec:skill_data_distribution}

\paraid{5.2.2-P1}\sentid{1}Figure~\ref{fig:data_scaling_families} reveals
several skill-level outcomes that are difficult to explain through apparent
motor difficulty alone.
\begin{itemize}[leftmargin=1.6em,itemsep=2pt,topsep=3pt]
  \item \sentid{2}\textbf{Demanding skills can emerge.} Stack requires precise
  placement and stability, Pass requires coordinated dual-arm transfer, and
  Straighten requires multi-stage reorientation with an angle-sensitive final
  pose; nevertheless, all three acquire meaningful capability as training
  scales.
  \item \sentid{3}\textbf{Similar motions, divergent outcomes.} Wipe and Sweep
  involve closely related tool-mediated surface motions but achieve
  substantially different SR.
\end{itemize}
\sentid{4}These contrasts motivate examining whether the uneven capability
distribution is related to skill-specific training coverage.

\paraid{5.2.2-P2}\sentid{1}Rather than relying on selected examples, we audit
training coverage across the complete 20-group atomic-skill taxonomy.
\sentid{2}We process the full multi-embodiment corpus using its step-captioned
instruction--action segments, convert their annotated frame spans to hours,
and map them to the same action-centric taxonomy as the evaluation.
\sentid{3}We exclude Close
from the quantitative relationship because its contact-based success criterion
makes it mechanically easier than the other groups.  \sentid{4}For each of the
remaining 19 groups, we pair its summed training hours with the SR of the G1-OP
30{,}000-hour model and analyze the relationship using log hours and logit SR.

\begin{figure}[H]
  \centering
  \includegraphics[width=\linewidth]{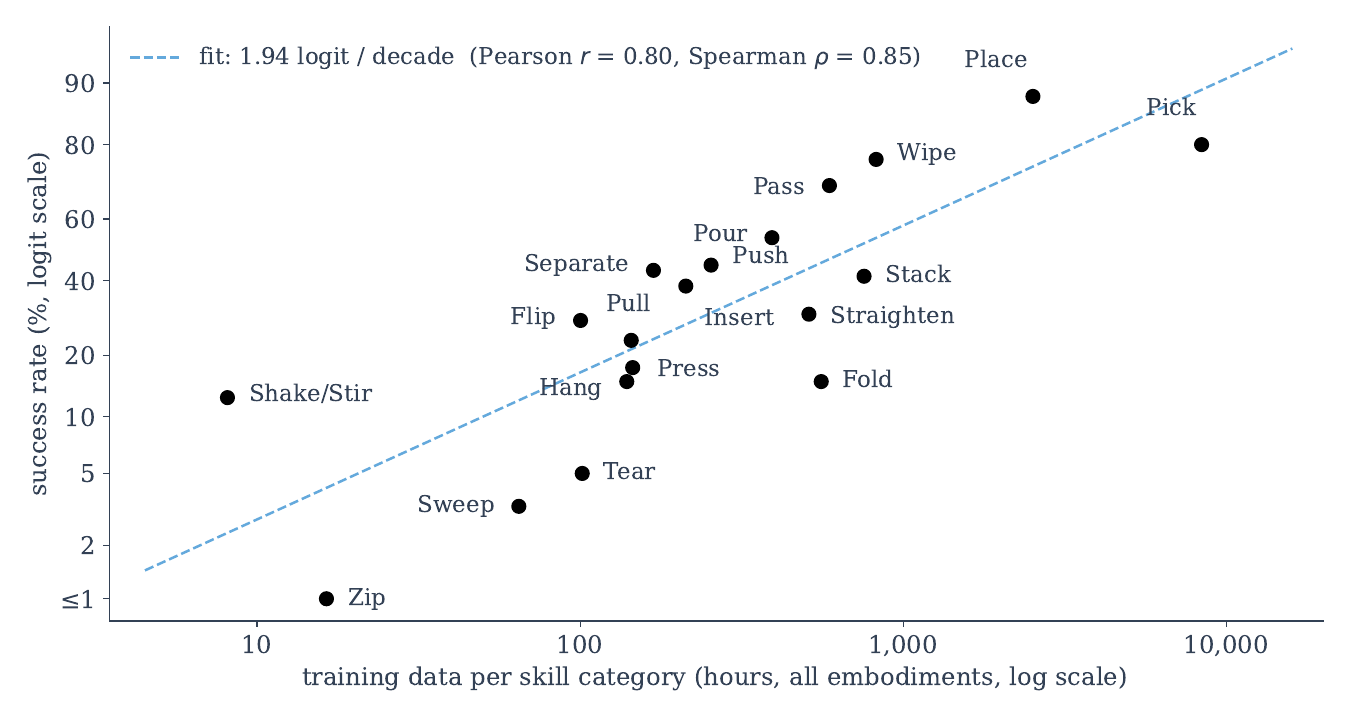}
  \caption{Skill-group training coverage and G1-OP zero-shot OOD success for
  the 30{,}000-hour model.  Each point is one of the 19 groups retained after
  omitting the mechanically trivial Close group; the
  horizontal axis reports the corresponding training hours across all
  embodiments on a logarithmic scale, and the vertical axis reports SR on a
  logit scale.  Coverage and success are strongly correlated
  (Pearson $r=0.80$; Spearman $\rho=0.85$), with a fitted gain of 1.94 logit
  units per decade of data.}
  \label{fig:skill_data_distribution}
\end{figure}

\paraid{5.2.2-P3}\sentid{1}Across the 19 nontrivial skill groups, training
coverage is strongly associated with zero-shot OOD success (Pearson $r=0.80$;
Spearman $\rho=0.85$).  \sentid{2}The fitted relationship corresponds to an increase of
1.94 logit units in SR for every tenfold increase in skill-specific training
hours.  \sentid{3}The qualitative irregularities visible in
Figure~\ref{fig:data_scaling_families} therefore form a systematic
distributional pattern rather than a collection of isolated cases.

\paraid{5.2.2-P4}\sentid{1}The fitted relationship provides a consistent
explanation for the qualitative observations above.
\begin{itemize}[leftmargin=1.6em,itemsep=2pt,topsep=3pt]
  \item \sentid{2}Despite their demanding execution, skills with substantial
  coverage achieve nontrivial performance: Pass receives 590.8 hours and
  improves from 34.0\% to 52.0\%, 60.0\%, and finally 70.0\%; Straighten
  receives 510.5 hours and remains at 0.0\% through 5{,}000 hours before
  reaching 30.0\% at 30{,}000 hours; and Stack receives 756.2 hours and rises
  from 21.4\% to 41.4\%.
  \item \sentid{3}Conversely, similar motions can lead to divergent outcomes:
  Wipe receives 824.1 hours and reaches 76.7\% SR, whereas Sweep receives 64.6
  hours and reaches only 3.3\%.
\end{itemize}

\paraid{5.2.2-P5}\sentid{1}Although intrinsic difficulty may explain part of
the cross-skill variation, the within-skill scaling trajectories in
Figure~\ref{fig:data_scaling_families} and the corpus-wide relationship in
Figure~\ref{fig:skill_data_distribution} make a difficulty-only explanation
insufficient.  \sentid{2}Taken together, these results provide converging
evidence that increasing both the scale and skill diversity of robot training
data---particularly the coverage of long-tail behaviors---should improve
pretrained zero-shot OOD capability across a broader range of manipulation
settings.

\subsection{Instruction Following in Complex Scenes}
\label{sec:language_following}

\paraid{5.3-P1}\sentid{1}Instruction following requires both selecting the
correct behavior among several visually plausible alternatives and remaining
responsive when the current instruction conflicts with an ongoing action or a
conventional scene relation.  \sentid{2}We evaluate these complementary aspects
in complex tabletop scenes using the final checkpoint of the full \methodname{} co-training
run on G1-OP, without further adaptation.  \sentid{3}Because GE-Act~2.0 remains
a low-level policy, these experiments test
instruction-dependent control over familiar executable behaviors rather than
high-level planning or open-ended reasoning.

\subsubsection{Fine-Grained Grounding across Instruction Dimensions}
\label{sec:language_dimensions}

\paraid{5.3.1-P1}\sentid{1}To minimize confounding from familiarity with the
underlying motor skill, execution is restricted to \emph{pick} and
\emph{place}, the two most frequent and reliable primitives in the robot data.
\sentid{2}A failure can therefore be interpreted with less ambiguity than on an
unfamiliar or long-horizon skill.

\begin{figure}[H]
  \centering
  \includegraphics[width=\linewidth]{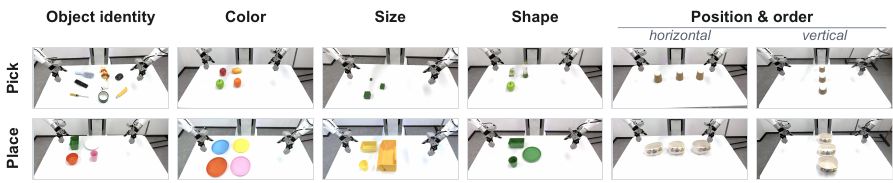}
  \caption{Controlled scenes for fine-grained instruction following, ordered
  as introduced in the text: object identity, color, size, shape, and
  position~\&~order.  Position and order share one setup, evaluated in a
  horizontal layout (left/right and ordinal references) and a vertical layout
  (near/far).  Top row: Pick scenes; bottom row: Place scenes.}
  \label{fig:language_scenes}
\end{figure}

\paraid{5.3.1-P2}\sentid{1}Within each primitive, we decompose instruction
grounding into six dimensions: object identity, color, size, shape,
position, and order.  \sentid{2}Size jointly
evaluates largest, smallest, medium, and related comparative expressions;
position covers direct left/right and near/far relations, whereas order covers
middle and ordinal references such as ``second from the
left.''  \sentid{3}The suite contains 33 Pick instructions and 26 Place
instructions, each evaluated over five trials, for a total of 295 real-robot
rollouts.  \sentid{4}Each dimension uses a controlled scene in which multiple
actions are visually feasible while only the instruction-specified target or
destination is correct.  \sentid{5}Figure~\ref{fig:language_scenes} shows the
paired Pick and Place setups for all five physical scene families; the complete
instruction list is provided in Appendix~\ref{app:language_details}.

\begin{figure}[H]
  \centering
  \includegraphics[width=\linewidth]{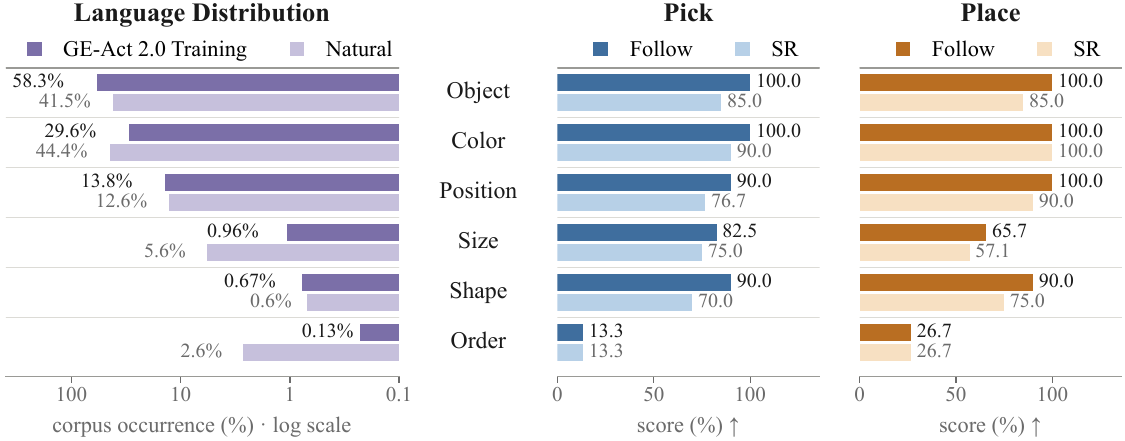}
  \caption{Fine-grained instruction following and language distribution across
  six grounding dimensions.  Left: occurrence in 95{,}010 human-written
  RefCOCOg expressions (Natural) and the GE-Act~2.0 training corpus, shown on a
  logarithmic scale.  Middle and right: Follow Score and full-task SR for Pick
  and Place, respectively.  Dimensions are ordered by their occurrence in the
  GE training corpus; descriptor categories are counted independently and may
  overlap.}
  \label{fig:language_following}
\end{figure}

\paraid{5.3.1-P3}\sentid{1}We report two complementary metrics: \emph{Follow
Score}, which measures whether the policy physically reaches only the
instruction-specified referent, and SR, which requires the full Pick or Place
behavior to be completed.  \sentid{2}For Pick, a trial satisfies Follow Score
when the gripper contacts the specified object without contacting another
object, even if it subsequently fails to lift the target.  \sentid{3}For
Place, Follow Score is satisfied when the carried object contacts the specified
target container and no other container, even if placement is not completed.
\sentid{4}The difference between Follow Score and SR therefore separates
incorrect instruction grounding from physical execution failure after the
correct referent has been reached.  \sentid{5}Figure~\ref{fig:language_following}
reports both metrics for Pick and Place over the six grounding dimensions and
places them beside the corresponding occurrence rates in natural referring
expressions and the GE-Act~2.0 training corpus; instruction-level results are
deferred to the appendix.

\paraid{5.3.1-P4}\sentid{1}GE-Act~2.0 reliably grounds object identity, color,
direct position, and shape, while size is less consistent and order remains
the primary weakness.  \sentid{2}Follow Score is at least 90.0\% for both Pick
and Place across those four dimensions, but decreases to 82.5\%/65.7\%
for size and 13.3\%/26.7\% for order.  \sentid{3}Across all 295 rollouts, the
overall Follow Score is 83.1\%, compared with a full-task SR of 72.9\%, showing
that some failures occur after the policy has already reached the correct
referent.

\paraid{5.3.1-P5}\sentid{1}This capability profile broadly mirrors the
long-tailed distribution of referring expressions: in both 95{,}010
human-written RefCOCOg expressions \citep{mao2016generation} and the GE-Act~2.0
training corpus, object and color form the head, position is intermediate, and
size, shape, and order are comparatively rare.  \sentid{2}The tail is
especially pronounced in GE training, where size, shape, and order occur in
only 0.96\%, 0.67\%, and 0.13\% of instructions, respectively.  \sentid{3}Order
is therefore both the least represented and weakest dimension, while size
shows a softer version of the same pattern.  \sentid{4}Shape is an informative
exception, achieving 90.0\% Follow Score for both primitives despite its low
occurrence, showing that frequency is not the only determinant of grounding
difficulty.  \sentid{5}Because frequency is not independently controlled and
the dimensions differ in perceptual and relational complexity, this experiment
establishes association rather than causation.  \sentid{6}Nevertheless, the
combined distribution and performance profile suggests that increasing the
diversity and volume of rare referring expressions---especially ordered and
comparative relations---is a promising route to stronger pretrained zero-shot
grounding; matching rules and complete counts are provided in
Appendix~\ref{app:language_details}.

\subsubsection{Instruction Following under Behavioral and Semantic Conflicts}
\label{sec:instruction_conflicts}

\paraid{5.3.2-P1}\sentid{1}We next examine two harder forms of instruction
conflict: behavioral conflict, where a revised command opposes a robot--scene
state already committed toward the previous behavior, and semantic conflict,
where the requested relation contradicts a conventional scene association.
\sentid{2}In preliminary experiments with earlier, weaker model variants, we
observed a characteristic failure mode: once the end effector had approached
an object or initiated an action, the policy often continued the visually
implied behavior even after the instruction changed.  \sentid{3}A plausible
explanation is a strong state--action prior in the training data: trajectories
in which the gripper has reached object A typically continue by grasping A
rather than redirecting to B.  \sentid{4}The behavioral-conflict tests
therefore ask whether the active instruction can override this state-conditioned
inertia.

\begin{figure}[H]
  \centering
  \includegraphics[width=\linewidth]{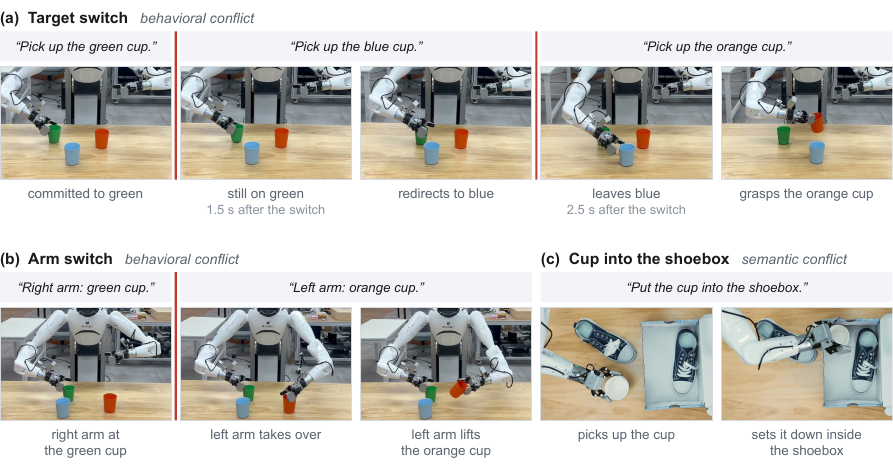}
  \caption{Instruction-conflict stress tests.  Each panel shows the active
  instruction, frames from one rollout, and the outcome; red rules mark where
  the instruction changed.  (a)~Target switch: the commanded cup changes from
  green to blue, then to orange, while the gripper is already at the previous
  target; the robot redirects each time and grasps the final target.
  (b)~Arm switch: the command moves from the right arm on the green cup to the
  left arm on the orange cup, and the left arm completes it.  (c)~Semantic
  conflict: the robot places a cup into a shoebox, following the explicit
  object--destination relation over the conventional scene association.
  Illustrative only; not part of the quantitative evaluation.}
  \label{fig:instruction_conflicts}
\end{figure}

\paraid{5.3.2-P2}\sentid{1}\emph{Behavioral conflicts.} We replace either the
target or the commanded arm only after the end effector executing the original
instruction is already near the target or has begun to grasp it, such that the
evolved physical state favors completing the previous behavior.
\begin{itemize}[leftmargin=1.6em,itemsep=2pt,topsep=3pt]
  \item \sentid{2}\emph{Target switch.} The commanded object changes from A to
  B, and later from B to C, each time when the end effector is already close
  to, or beginning to grasp, the previous target
  (Figure~\ref{fig:instruction_conflicts}a).
  Immediately after each switch, the robot retains brief residual motion toward
  the previous target, but subsequently redirects its trajectory and finally
  grasps C.
  \item \sentid{3}\emph{Arm switch.} The command changes from the right arm
  on the green cup to the left arm on the orange cup when the right end effector
  is already at the green cup (Figure~\ref{fig:instruction_conflicts}b).  The
  robot disengages the right arm and completes the new command with the left
  arm with little visible hesitation.
\end{itemize}

\paraid{5.3.2-P3}\sentid{1}\emph{Semantic conflicts.} We retain familiar
objects and motor primitives but request a physically feasible relation that
conflicts with the conventional object relations implied by the scene.
\sentid{2}This construction tests whether behavior is controlled by the
explicit instruction or by familiar visual--action associations memorized from
training.  \sentid{3}For example, we place a cup beside a shoe and a shoebox, then
instruct the robot to put the cup into the shoebox
(Figure~\ref{fig:instruction_conflicts}c).  \sentid{4}The policy
completes the unusual placement, following the explicit object--destination
relation rather than defaulting to the more conventional scene association.

\paraid{5.3.2-P4}\sentid{1}The demonstrations reveal distinct qualitative
dynamics: arm designation can be revised with little hesitation, whereas a
target change near contact requires recovery from the previous trajectory.
\sentid{2}Successful counter-conventional placement further provides
qualitative evidence that an explicit object--destination command can override
a familiar scene prior.  \sentid{3}These cases complement the static dimension-level
scores by exposing how instruction-dependent control behaves under temporal
and semantic conflict.

\subsection{OOD Performance on Simulation Benchmarks}
\label{sec:simulation_benchmarks}

\paraid{5.4-P1}\sentid{1}Real-robot evaluation is our primary measure of a
pretrained policy's zero-shot capability, but matching hardware, scenes, and
evaluation conditions across independently developed models is difficult.
\sentid{2}We therefore complement the main real-robot experiments with three
standardized simulation benchmarks spanning environment randomization,
instruction following, and controlled visual and embodiment perturbations.
\sentid{3}These benchmarks cannot directly establish out-of-the-box real-world
generalization: the simulator-to-real gap is substantial, and each benchmark
includes an in-distribution training or adaptation stage.  \sentid{4}Because
GE-Act~2.0 is pretrained mostly on real-world data, we first SFT it on each
benchmark's prescribed in-distribution training condition, putting it on the
same footing as models adapted specifically to that simulator, and then
evaluate only held-out OOD simulation conditions.  \sentid{5}We deliberately
do not develop a benchmark-specific post-training strategy.

\paraid{5.4-P2}\sentid{1}RoboTwin Clean-to-Random~\citep{chen2025robotwin2}
trains on the clean setting and tests under domain randomization;
GenieSim-Instruction~\citep{yin2026geniesim} trains
on its released instruction-following data and tests unseen scene
configurations; and LIBERO-Plus~\citep{fei2025liberoplus} adapts on standard
LIBERO and evaluates camera, robot-state, language, lighting, background,
sensor-noise, and layout perturbations.  \sentid{2}Despite this deliberately
simple adaptation, GE-Act~2.0 is the strongest compared model on RoboTwin and
GenieSim-Instruction, and remains competitive on LIBERO-Plus.  \sentid{3}Full
protocols, comparisons, and per-axis results are provided in
Appendix~\ref{app:simulation}.
\section{Conclusion}
\label{sec:conclusion}

We introduced GE-Act~2.0, a world--action model designed to study the
pretraining and scaling of visual generation and action prediction as a unified
system. GE-Act~2.0 combines a control-oriented autoencoder that preserves
control-relevant information, a single-step visual planner that enables
decoupled pretraining, and \methodname{}, which aligns generated futures with
compatible action supervision during co-training.

Across zero-shot OOD evaluations on 100 real-robot tasks and two embodiments,
increasing manipulation data produces broad improvements across nearly all
evaluated skill groups. Performance also improves on the sparsely represented
embodiment, suggesting useful transfer from the broader training corpus, while
skill-specific data coverage strongly predicts downstream success.
Complementary experiments demonstrate fine-grained instruction grounding and
adherence to explicit commands under challenging behavioral and semantic
conflicts. Together, these findings establish GE-Act~2.0 as a scalable approach
to learning transferable manipulation capabilities through world--action
pretraining.

\section{Limitations and Outlook}
\label{sec:limitations}

GE-Act~2.0 opens a direct route for learning from egocentric manipulation video,
but the present study has not yet explored this source at the scale its
availability permits. We view egocentric video as a particularly promising
source of diverse physical interactions and task semantics, and preliminary
signals from our training mixtures are encouraging. A key direction for future
work is therefore to scale egocentric-video pretraining and rigorously establish
how its volume, diversity, and composition affect visual-generation pretraining and downstream
robot capability.

GE-Act~2.0 currently focuses on System-1-style instruction-conditioned control,
mapping the current observation and language instruction directly to low-level
manipulation without explicit deliberation. Robust task execution in open
environments will additionally require complementary System-2 capabilities for
deliberate reasoning, long-horizon
planning, task decomposition, memory, and self-correction. We believe that
coupling a scalable world--action model with such a deliberative system is a
promising path toward general-purpose embodied intelligence, and future work
will study how these two levels can plan, act, and improve together in closed
loop.

\clearpage
\section{Contributions and Acknowledgments}
\label{sec:contributions_acknowledgments}

\vspace{0.4em}
\noindent{\small\bfseries Core Contributors}
\par\vspace{0.15em}
\begingroup
\small
\setlength{\tabcolsep}{0pt}
\renewcommand{\arraystretch}{1.15}
\begin{tabular}{@{}*{5}{>{\raggedright\arraybackslash}p{0.2\textwidth}}@{}}
Renhang Liu    & Wenzhi Zhao & Zhuo Yang    & Liliang Chen  & Pengfei Zhou \\
Shengcong Chen & Guanghui Ren &             &                &              \\
\end{tabular}
\endgroup

\medskip
\noindent{\small\bfseries Contributors}
\par\vspace{0.15em}
\begingroup
\small
\setlength{\tabcolsep}{0pt}
\renewcommand{\arraystretch}{1.15}
\begin{tabular}{@{}*{5}{>{\raggedright\arraybackslash}p{0.2\textwidth}}@{}}
Youlun Peng     & Rongjun Jin      & Nan Wang      & Sukai Wang       & Xindong He \\
Jinyuan Feng    & Ziyu Xiong      & Linqing Zhong & Yifei Wei        & Feng Han \\
Long Zhang      & Da Huang        & Nanshu Zhao   & Chenghao Yin     & Mo Wu \\
Zhaodong Yan    & Kongtao Hu      & Yuxiang Yan   & Aogelijiang Niyazi & Yu Fang \\
Jia Zeng        & Lizhu Meng      & Daizhen Lv    & Haoyu Cao        & Zhiwen Hou \\
Lianjin Ye      & Yuehan Niu      & Zhikai Cai    & Xuan Hu          & Hui Min \\
Xiongfeng Cai   & Yue Liao        & Jing Wu       &                  & \\
\end{tabular}
\endgroup

\medskip
\noindent{\small\bfseries Academic Advisors}
\par\vspace{0.15em}
\begingroup
\small
\setlength{\tabcolsep}{0pt}
\begin{tabular}{@{}*{5}{>{\raggedright\arraybackslash}p{0.2\textwidth}}@{}}
Soujanya Poria & Ye Li & Sanping Zhou & & \\
\end{tabular}
\endgroup

\medskip
\noindent{\small\bfseries Project Leaders}
\par\vspace{0.15em}
\begingroup
\small
\setlength{\tabcolsep}{0pt}
\begin{tabular}{@{}*{5}{>{\raggedright\arraybackslash}p{0.2\textwidth}}@{}}
Liliang Chen & Guanghui Ren & & & \\
\end{tabular}
\endgroup

\medskip
\noindent{\small\bfseries Supervisors}
\par\vspace{0.15em}
\begingroup
\small
\setlength{\tabcolsep}{0pt}
\begin{tabular}{@{}*{5}{>{\raggedright\arraybackslash}p{0.2\textwidth}}@{}}
Guanghui Ren & Maoqing Yao & & & \\
\end{tabular}
\endgroup

\medskip
\noindent\begin{minipage}{\textwidth}
{\small\bfseries Acknowledgments}
\par\vspace{0.15em}
We appreciate Maniformer's R\&D team for their in-depth technical support on
data. We thank Yifei Chen for helpful discussions and assistance with polishing
the paper.
\end{minipage}

\bibliography{refs}

\appendix
\section{OOD Generalization in Simulation}
\label{app:simulation}

\paraid{AppSim-P1}\sentid{1}Real-robot evaluation is central to this report,
but matching hardware, scenes, and evaluation conditions across independently
developed models is difficult.  \sentid{2}We therefore complement the main
real-robot experiments with three standardized simulation benchmarks spanning
environment randomization, instruction following, and controlled visual and
embodiment perturbations.  \sentid{3}Because GE-Act~2.0 is pretrained mostly
on real-world data, we first SFT it on each benchmark's designated
in-distribution training condition, placing it on the same footing as models
adapted specifically to that simulator, and evaluate it only on held-out OOD
simulation conditions.  \sentid{4}This protocol provides a more comparable
cross-model evaluation while preserving a separation between adaptation and
OOD evaluation, following the principle used by
Qwen-RobotManip~\citep{qwenrobotmanip2026}.

\subsection{RoboTwin Clean-to-Random Generalization}
\label{app:simulation:robotwin}

\paraid{AppSim-RT-P1}\sentid{1}RoboTwin Clean-to-Random tests whether a policy
adapted only to a clean simulator remains effective as the scene distribution
changes~\citep{chen2025robotwin2}.  \sentid{2}We SFT GE-Act~2.0 on the
clean training setting and evaluate it without further adaptation both on the
clean setting (Clean-to-Clean) and under the benchmark's domain randomization
(Clean-to-Random).  \sentid{3}Table~\ref{tab:robotwin_clean2rand}
reports the task-averaged success rate for both settings.
\sentid{4}The compared models are StarVLA~\citep{starvla2026},
GR00T~N1.7~\citep{nvidia2025gr00t}, X-VLA~\citep{zheng2025xvla},
$\pi_{0.5}$~\citep{intelligence2025pi05},
GigaBrain-0.7~\citep{gigabrain2026}, and
Qwen-RobotManip~\citep[Qwen-RM;][]{qwenrobotmanip2026}, with baseline results taken from
the official RoboTwin~2.0 leaderboard where available.


\begin{table}[H]
\captionsetup{skip=7pt}
\caption{RoboTwin~2.0 success rate (\%, $\uparrow$).  All models are adapted
on 50 clean demonstrations per task and evaluated over 100 trials per task on
the clean (Clean-to-Clean) and domain-randomized (Clean-to-Random) settings.
Baseline results are from the official RoboTwin~2.0 leaderboard (snapshot of
2026-10-24) and, for GR00T~N1.7 and Qwen-RM, from the Qwen-RM technical
report.  The best
result in each row is bolded.}
\label{tab:robotwin_clean2rand}
\centering
\tablebodyfont
\setlength{\tabcolsep}{5pt}
\begin{tabular}{lrrrrrr>{\columncolor{RedOrange!8}}r}
\toprule
Setting & StarVLA & GR00T-N1.7 & X-VLA & $\pi_{0.5}$ & GigaBrain-0.7 & Qwen-RM & \textbf{GE-Act~2.0} \\
\midrule
Clean-to-Clean & 46.52 & 43.60 & 67.96 & 70.66 & 66.80 & 84.70 & \textbf{88.54} \\
Clean-to-Random & 3.16 & 20.70 & 20.95 & 46.00 & 67.92 & 69.40 & \textbf{81.48} \\
\bottomrule
\end{tabular}
\end{table}

\paraid{AppSim-RT-P2}\sentid{1}GE-Act~2.0 achieves the strongest performance
in both settings.  \sentid{2}Under domain randomization, it reaches 81.48\%
success, 12.08 percentage points above the strongest baseline,
Qwen-RM at 69.40\%.  \sentid{3}Its performance decreases by only
7.06 points from Clean-to-Clean to Clean-to-Random, whereas Qwen-RM
decreases by 15.30 points, $\pi_{0.5}$ by 24.66, and StarVLA by 43.36,
indicating substantially better retention under scene randomization.

\subsection{GenieSim-Instruction: OOD Instruction Following}
\label{app:simulation:geniesim}

\paraid{AppSim-GS-P1}\sentid{1}GenieSim-Instruction provides an OOD test of
whether a policy can select the correct manipulation behavior from diverse
language specifications~\citep{yin2026geniesim}.  \sentid{2}The benchmark
itself separates training from evaluation: we SFT on its released
instruction-following training demonstrations and evaluate on prescribed
configurations with unseen background textures, object positions, and related
scene variations.  \sentid{3}The ten tasks probe color, number, shape, size,
object type, specific-object reference, common-sense reference, logical
composition, and object straightening.  \sentid{4}We report the official
normalized task score and its average across tasks in
Table~\ref{tab:geniesim_instruction}.
\sentid{5}The compared models are $\pi_0$~\citep{black2024pi0}, GR00T~N1.7~\citep{nvidia2025gr00t}, LingBot-VLA~2.0~\citep{wu2026lingbotvla2}, $\pi_{0.5}$~\citep{intelligence2025pi05}, and ACoT-VLA~\citep{zhong2026acotvla}.

\begin{table}[H]
\captionsetup{skip=7pt}
\caption{Results on the OOD GenieSim-Instruction benchmark (normalized score,
$\uparrow$).  Evaluation configurations vary background textures, object
positions, and other scene factors relative to the training demonstrations.
The best result in each row is bolded.}
\label{tab:geniesim_instruction}
\centering
\tablebodyfont
\setlength{\tabcolsep}{3.6pt}
\begin{tabular}{lrrrrr>{\columncolor{RedOrange!8}}r}
\toprule
Task & $\pi_0$ & GR00T-N1.7 & LingBot-VLA~2.0 & $\pi_{0.5}$ &
ACoT-VLA & \textbf{GE-Act~2.0} \\
\midrule
pick\_billiards\_color & 0.50 & 0.73 & 0.88 & 0.83 & 0.91 & \textbf{0.95} \\
pick\_block\_color & 0.49 & 0.80 & 0.80 & 0.87 & \textbf{0.96} & 0.92 \\
pick\_block\_number & 0.40 & 0.61 & 0.72 & \textbf{0.82} & 0.76 & 0.71 \\
pick\_block\_shape & 0.20 & 0.44 & 0.43 & \textbf{0.55} & 0.52 & 0.52 \\
pick\_block\_size & 0.34 & 0.62 & 0.68 & 0.79 & \textbf{0.83} & 0.76 \\
pick\_common\_sense & 0.19 & 0.68 & 0.63 & 0.59 & 0.66 & \textbf{0.74} \\
pick\_follow\_logic\_or & 0.39 & 0.72 & 0.69 & 0.81 & 0.79 & \textbf{0.89} \\
pick\_object\_type & 0.42 & 0.72 & \textbf{0.82} & 0.80 & 0.80 & 0.78 \\
pick\_specific\_object & 0.37 & 0.69 & 0.87 & 0.81 & 0.83 & \textbf{0.89} \\
straighten\_object & 0.38 & 0.45 & 0.56 & \textbf{0.59} & 0.51 & 0.54 \\
\midrule
Average & 0.368 & 0.646 & 0.708 & 0.746 & 0.757 & \textbf{0.770} \\
\bottomrule
\end{tabular}
\end{table}

\paraid{AppSim-GS-P2}\sentid{1}GE-Act~2.0 obtains the highest average score
of 0.770, compared with 0.757 for ACoT-VLA and 0.746 for $\pi_{0.5}$.
\sentid{2}It leads on four tasks, with its clearest margins on logical
composition and common-sense reference, where it improves over the previous
best by 0.08 and 0.06, respectively.  \sentid{3}The advantage is not uniform:
the strongest baseline remains ahead on number, shape, size, and several
direct attribute-matching tasks.  \sentid{4}The aggregate improvement
therefore reflects complementary instruction-following strengths rather than
uniform dominance across every language category.

\subsection{LIBERO-Plus Robustness across Perturbation Axes}
\label{app:simulation:liberoplus}

\paraid{AppSim-LP-P1}\sentid{1}LIBERO-Plus tests whether a policy adapted on
standard LIBERO remains reliable under controlled changes to its observations,
instructions, and execution conditions~\citep{fei2025liberoplus}.
\sentid{2}We SFT GE-Act~2.0 on the standard LIBERO training data and evaluate
it on seven LIBERO-Plus perturbation axes: camera viewpoint, robot initial
state, language, lighting, background, sensor noise, and object layout.
\sentid{3}Table~\ref{tab:liberoplus} reports the success rate for each axis and
the overall benchmark average.
\sentid{4}The compared models are StarVLA~\citep{starvla2026} and $\pi_{0.5}$~\citep{intelligence2025pi05}.

\begin{table}[H]
\captionsetup{skip=7pt}
\caption{LIBERO-Plus success rate (\%, $\uparrow$) after adaptation on standard
LIBERO.  Each column evaluates a held-out perturbation axis.  Overall is the
success rate over all 10{,}030 task instances, so axes are weighted by their
instance counts, following the benchmark's protocol; all models are scored
the same way.  The best result in each column is bolded.}
\label{tab:liberoplus}
\centering
\tablebodyfont
\setlength{\tabcolsep}{5.2pt}
\begin{tabular}{lrrrrrrrr}
\toprule
Model & Camera & Robot & Language & Light & Background & Noise & Layout & Overall \\
\midrule
StarVLA & 52.5 & 49.8 & \textbf{88.5} & 95.7 & \textbf{95.7} & 73.0 & 76.9 & 74.1 \\
$\pi_{0.5}$ & 78.4 & \textbf{73.6} & 80.8 & \textbf{96.2} & 94.1 & 89.0 & \textbf{84.5} & \textbf{84.4} \\
\rowcolor{RedOrange!8}
\textbf{GE-Act~2.0} & \textbf{94.1} & 50.7 & 81.4 & 94.0 & 60.5 & \textbf{95.5} & 83.1 & 80.4 \\
\bottomrule
\end{tabular}
\end{table}

\paraid{AppSim-LP-P2}\sentid{1}GE-Act~2.0 reaches 80.4\% overall, exceeding
StarVLA by 6.3 percentage points while remaining 4.0 points below
$\pi_{0.5}$.  \sentid{2}It performs best under camera and sensor-noise
perturbations, reaching 94.1\% and 95.5\%, respectively, and remains close to
$\pi_{0.5}$ under language and layout changes.  \sentid{3}Robot-state and
especially background perturbations remain the principal weaknesses, showing
that robustness is not yet uniform across the seven axes.

\paraid{AppSim-P2}\sentid{1}Taken together, the simulation results show the
strongest generalization under compounded scene randomization and OOD
instruction following.  \sentid{2}The per-axis LIBERO-Plus results provide a
more diagnostic picture, identifying background and robot-state shifts as
important remaining failure modes.  \sentid{3}These standardized comparisons
therefore complement the real-robot atomic-task evaluation by exposing both
broad OOD capability and benchmark-specific limitations.

\section{Training Recipes and Hyperparameters}
\label{app:recipes}



Unless stated otherwise, all stages use bf16 mixed precision with DeepSpeed
ZeRO stage~2, AdamW with $(\beta_1,\beta_2)=(0.9,0.95)$ and weight decay
$10^{-5}$, gradient-norm clipping at $1.0$, a constant learning rate after a
linear warmup, and no classifier-free guidance or caption dropout.  Batch
sizes are per GPU.

\subsection{Architectures}
\label{app:recipes:arch}

\begin{table}[H]
\caption{Trained components.  Parameter counts are read from the final
co-trained checkpoint; the frozen Qwen3.5-2B VLM (2.72\,B parameters) is
excluded.}
\label{tab:recipes_arch}
\centering
\tablebodyfont
\setlength{\tabcolsep}{5pt}
\begin{tabular}{@{}lp{0.74\linewidth}@{}}
\toprule
Component & Configuration \\
\midrule
\coaename{} &
Framewise deep-compression autoencoder (DC-AE layout) with $64\times$ spatial
downsampling and 512 latent channels; a $256\times384$ frame gives a
$4\times6$ grid of 24 tokens.  Latents are standardized per channel with
corpus statistics. \\
\addlinespace[3pt]
\svpname{} flow generator &
2.51\,B parameters.  A 36-block diffusion transformer of width 2048 (32
heads): 20 shared blocks followed by 8 blocks for the mean-velocity head
$u_\theta$ and 8 parallel blocks for the instantaneous-velocity head
$v_\theta$.  Every block cross-attends to the fused VLM states; every third
block attends across views.  Conditioning enters through adaptive layer
normalization as the sum of sinusoidal embeddings of $t$ and of $r$ and a
learned view-identity embedding.  Per view the input is one conditioning
frame plus six predicted frames, i.e., $7\times24=168$ tokens. \\
\addlinespace[3pt]
VLM (frozen) &
Qwen3.5-2B.  The prompt holds the current head-view frame (96 vision tokens)
and the instruction in a fixed window of 192 tokens; the text-span states of
all 25 layers are fused by the layer gate of Eq.~\eqref{eq:vlmgate}. \\
\addlinespace[3pt]
IDM &
0.56\,B parameters.  28 transformer blocks of width 1152 (16 heads) with
self-attention over action tokens and cross-attention to the visual latent
tokens of the current and predicted frames.  The flow time and a learned
embodiment embedding enter through adaptive layer normalization; actions are
32-dimensional joint-space vectors, zero-padded across embodiments, and the
proprioceptive state enters through a small MLP. \\
\bottomrule
\end{tabular}
\end{table}

\subsection{Temporal Schedule and Sample Construction}
\label{app:recipes:schedule}

\paragraph{Frames and actions.}
The defaults of Eq.~\eqref{eq:schedule} are $(N_d,N_s,H)=(4,2,52)$ frames at
30\,Hz, to which all sources are aligned, so $H$ covers $1.73$\,s.  The clip
end $f_0+T$ is the end of the current step-caption segment, with a small
jitter and an occasional extension to the end of the following segment.  Each
sample carries 52 dense actions at the control rate and two sparse actions at
the sparse frames; actions are absolute joint positions standardized per
source, and sparse action tokens receive a loss weight of $0.1$.  Frames are
resized to $256\times384$.

\subsection{\svpname{} Pretraining}
\label{app:recipes:svp}

The flow time is $t=s(\tau)$ with $\tau\sim\sigma(\mathcal{N}(0,1))$ and the
shift $s(\tau)=2.7\tau/(1+1.7\tau)$; the endpoint is $r=s(\rho\tau)$ with
$\rho\sim\mathcal{U}(0,1)$.  In $75\%$ of the samples $r=t$, which reduces
the objective to flow matching; the rest train the interval $[r,t]$.  The
conditioning frame's latent is mixed with Gaussian noise at a level drawn
uniformly in $[0,0.1]$.  The Jacobian--vector product of Eq.~\eqref{eq:compound}
is computed exactly with forward-mode autodiff, and the detached correction
$(t-r)\,\mathrm{d}u_\theta/\mathrm{d}t$ is clamped element-wise to $[-5,5]$;
the time embedders are zero-initialized so that the correction starts at
zero.  The losses of Eq.~\eqref{eq:loss} are averaged over latent elements
with dense and sparse frames weighted equally, and each per-sample loss is
normalized as $\mathcal{L}/\operatorname{sg}[\mathcal{L}+0.1]$ separately for
the two heads.  In pretraining only, $5\%$ of the samples replace the
instruction with a clean goal frame and learned null-text tokens.

The \svpname{} is trained from random initialization with learning rate
$10^{-4}$ (20,000-step warmup) and a per-GPU batch of 8 and later 12, with
the data sources of Figure~\ref{fig:data_mixtures} introduced
progressively; the learning rate is decayed linearly
to zero at the end of pretraining.

\subsection{IDM Pretraining}
\label{app:recipes:idm}

The IDM is trained on recorded futures with Eq.~\eqref{eq:idm}, a
logit-normal flow time $t\sim\sigma(\mathcal{N}(0,1))$ with unit loss
weight, and light Gaussian blur and noise on the visual latent tokens.  It
is trained with a per-GPU batch of 16 and learning rate $10^{-4}$ with a
5,000-step warmup on the \idmPretrainHours{}-hour mixture of
Figure~\ref{fig:data_mixtures}.  The controlled ablation of
Section~\ref{sec:evidence} uses the same IDM checkpoint.

\subsection{\methodname{} Co-training}
\label{app:recipes:kaso}

\paragraph{Fused update.}
Each optimizer step draws one batch of instruction--video--action samples
and accumulates three micro-steps on it: $\mathcal{L}_{\mathrm{SVP}}$ on the
recorded video, $\mathcal{L}_{\mathrm{IDM}}$ on the recorded future, and
$\mathcal{L}^{\mathrm{sel}}_{\mathrm{E2E}}$ on the selected generated future,
with coefficients
$(\lambda_{\mathrm{SVP}},\lambda_{\mathrm{IDM}},\lambda_{\mathrm{E2E}})=(1,0.1,1)$
and the accumulated gradient divided by three.  Every sample thus contributes
to the recorded and the generated action loss once per update (a 1:1 mixing
ratio).  The flow generator and the IDM are fully trainable; the VLM and
\coaename{} stay frozen.

\paragraph{Selection.}
The group size is $N=4$ with $k=1$ and the probe level is $t_p=0.95$; the
energy of Eq.~\eqref{eq:energy} is averaged over the 52 dense action tokens
with the \coaename{}-encoded recorded future as reference.  Selection is a
plain best-of-$N$ choice: the single lowest-energy candidate receives the
generated-future action loss with unit weight, and the remaining candidates
are discarded.  The selected loss uses a freshly drawn flow time and action
noise.

\paragraph{Runs.}
All ladder rungs start from the same pretrained \svpname{} and IDM and use
learning rate $10^{-4}$ with a 5,000-step warmup and a per-GPU batch of 16 on
nested slices of the co-training corpus; the largest rung uses the full corpus
of Figure~\ref{fig:data_mixtures}.

For the controlled ablation of Section~\ref{sec:evidence}, all arms use the
300-hour mixture, the same pretrained components, and a per-GPU batch of 8.
\methodname{} was trained with learning rate $10^{-4}$ and a 1,000-step
warmup; E2E+PT uses the same fused update with selection disabled ($k=N$),
and E2E uses $\mathcal{L}_{\mathrm{E2E}}$ alone.

\subsection{Deployment}
\label{app:recipes:deploy}

At deployment the policy receives the three camera views at $256\times384$,
the proprioceptive state, and the instruction.  The \svpname{} produces the
future latents in one forward pass at $(r,t)=(0,1)$, and the IDM integrates
its action flow with five Euler steps without guidance.  The first 30 actions
of the dense chunk are executed at the control rate before the next
prediction.  All results in
Section~\ref{sec:scaling} use the last checkpoint of the corresponding run
with these fixed settings and no per-task checkpoint selection; the reset and scoring protocol is given in
Appendix~\ref{app:full_results}.

\subsection{Toy Two-Stage System}
\label{app:toy}

\paragraph{Data-generating process.}
The toy experiment uses an unconditional two-stage mapping
$\epsilon\mapsto\hat z\mapsto\hat a$ with $z\in\mathbb{R}^3$ and
$a\in\mathbb{R}$.  The $(z_1,z_2)$ marginal is an equal-weight mixture of four
Gaussians with means $\mu_{ij}=(1.5i,1.8j)$ for
$i,j\in\{-1,1\}$ and per-axis standard deviations $0.04$ and $0.18$.
The third coordinate is exactly zero for every recorded sample.  We verify at
runtime that the largest empirical deviation of a component weight from
$0.25$ is below $0.02$.  The action is deterministic,
$a=z_1z_2/(1.5\times1.8)$, so the four video modes map to two balanced action
modes at $a=\pm1$.

\paragraph{Networks and initialization.}
The generator $A$ is a time-conditioned velocity field implemented as a tanh
MLP with widths $4$--$24$--$24$--$3$ and input $[x,t]$.  Its hidden layers use
Xavier-uniform initialization with gain $0.65$, its output layer uses gain
$0.12$, and all biases are zero.  The deterministic IDM proxy $B$ is a tanh
MLP with widths $3$--$24$--$24$--$1$; its hidden and output Xavier gains are
$0.80$ and $0.50$, respectively, with zero biases.  Before entering $B$, each
video coordinate is divided by $(1.5,1.8,s_z)$ with $s_z=0.2$.  All four arms
start from identical copies of $A$ and $B$; a runtime check confirms that their
step-0 outputs agree exactly.  The point-estimate IDM exposes conditional-mean
regression directly while retaining the video--action compatibility problem
studied in the full system.

\paragraph{Shared optimization.}
For flow matching, we draw $t\sim\mathcal{U}(0,1)$ and use the linear bridge
$x_t=(1-t)\epsilon+t z$ with velocity target $z-\epsilon$.  Generated samples
are obtained with 48 left-endpoint Euler steps, evaluating the velocity at
$t_i=i/48$.  Every arm uses AdamW for 7,000 steps with learning rate
$2\times10^{-3}$, weight decay $10^{-5}$, batch size 256, and global
gradient-norm clipping at 5.  The learning rate is halved every 650 steps
starting at step 1,900.  The pipeline weight $\rho_s$ is zero through step 300,
increases linearly to one over the next 300 steps, and remains one thereafter.

Let $\mathcal{L}_{\mathrm{FM}}$ denote the video flow-matching loss,
$\mathcal{L}_{\mathrm{real}}=\operatorname{MSE}(B(z),a)$ the IDM loss on
recorded pairs, and
$\mathcal{L}_{\mathrm{pipe}}=\operatorname{MSE}(B(A(\epsilon)),a)$ the
generated-video action loss.  The four objectives are
\begin{align*}
\text{Decoupled Pretraining:}\quad
  & \mathcal{L}_{\mathrm{FM}}+\mathcal{L}_{\mathrm{real}}, \\
\text{End-to-End Co-training:}\quad
  & \rho_s\,\mathcal{L}_{\mathrm{pipe}}, \\
\text{E2E + Pretraining Losses:}\quad
  & 0.3\,\mathcal{L}_{\mathrm{FM}}+0.1\,\mathcal{L}_{\mathrm{real}}
    +\rho_s\,\mathcal{L}_{\mathrm{pipe}}, \\
\text{\methodname{}:}\quad
  & 0.3\,\mathcal{L}_{\mathrm{FM}}+0.1\,\mathcal{L}_{\mathrm{real}}
    +\rho_s\,\mathcal{L}_{\mathrm{sel}}^{(8)},
\end{align*}
where $\mathcal{L}_{\mathrm{sel}}^{(8)}$ is the pipeline loss restricted to the
selected candidate, defined below.  Decoupled Pretraining keeps unit weights on
both pretraining losses, whereas the connected arms reduce them to $0.3$ and
$0.1$; this weighting favors the baseline.

\paragraph{Selection and replay.}
For each batch row, the toy \methodname{} arm draws eight generator noises and
integrates all eight candidates without gradients.  It scores candidate $k$ by
$\bigl(B(\hat z_k)-a\bigr)^2$, selects the minimum-energy candidate, and then
re-integrates from that candidate's retained noise with gradients enabled.
Only this replayed trajectory contributes
$\mathcal{L}_{\mathrm{sel}}^{(8)}$.  We verify that the replayed trajectory
matches the selected no-gradient trajectory exactly.  The eight-way selection
is training-only: evaluation draws one generator sample per input for every
arm.

\paragraph{Evaluation and stress-test controls.}
Figure~\ref{fig:toy_cascade} uses seed~0 and evaluates all arms on the same
fixed batch of 12,000 Gaussian noises.  Action densities are 220-bin normalized
histograms on $[-1.7,1.7]$, smoothed by a discrete Gaussian with standard
deviation 1.6 bins (approximately $0.025$ action units); all panels share the
same vertical scale.

The choice $s_z=0.2$, rather than the default value 1, creates a controlled
off-manifold sensitivity stress test by amplifying the IDM's response to the
third coordinate fivefold.  It is shared by all four arms and does not change
any recorded input because recorded data satisfy $z_3=0$.  At seed~0, the IDM
RMSE on recorded pairs is $0.0112$ for both $s_z=0.2$ and $s_z=1.0$.
Furthermore, the Decoupled arm has the lowest recorded-pair IDM error and the
lowest video sliced-$W_1$ distance among the learned arms
($0.068$, versus $0.103$ for \methodname{}), but fails when its IDM is applied
to generated off-manifold video.

The displayed figure is one run from a nontrivial seed spread.  Across seeds
0--4, the maximum action-density height of Decoupled Pretraining relative to
the ground-truth maximum is $0.240$, $0.617$, $0.647$, $0.955$, and $1.035$,
respectively.  We therefore interpret Figure~\ref{fig:toy_cascade} as a
controlled illustration of the off-manifold failure mechanism rather than as
a seed-averaged performance benchmark.

\section{Full Results}
\label{app:full_results}

\subsection{Atomic-Skill Taxonomy and Evaluation Details}
\label{app:task_taxonomy}

Our real-robot evaluation suite contains 100 atomic task instances in 20 skill
groups (Table~\ref{tab:atomic_taxonomy}).  Each task is evaluated on both
G1-OP and G2-90D using the same instruction and an observable,
instruction-specific terminal success criterion fixed before evaluation.  For
example, Pick requires a successful grasp and lift of the specified object,
Place requires the object to reach the instructed receptacle, and Close
requires the target to be fully closed.  The analogous terminal condition is
used for each remaining action type.

For every task, embodiment, and training-data scale, we run 10 trials under a
standardized reset procedure.  Within each task and embodiment, the reset
configurations are matched across checkpoints; scenes, backgrounds, lighting,
and test object instances are excluded from pretraining and co-training.  We
report empirical success rate (SR), i.e., the percentage of successful trials.
The task names below are compact descriptions; the evaluation uses the full
instruction wording and the indicated arm configuration.

\begin{longtable}{@{}p{0.90in}p{5.15in}@{}}
\caption{Atomic-task taxonomy for the 100-task real-robot zero-shot OOD suite.
Parenthetical counts sum to 100.  Unless an arm or both arms are explicitly
specified, the task is executed with the right arm.}\label{tab:atomic_taxonomy}\\
\toprule
Skill group & Atomic task instances \\
\midrule
\endfirsthead
\caption[]{Atomic-task taxonomy for the 100-task real-robot zero-shot OOD suite (continued).}\\
\toprule
Skill group & Atomic task instances \\
\midrule
\endhead
\bottomrule
\endfoot
Pick (13) & Red apple; toothpaste; red cube; water bottle (two scene variants);
marker pen from water cup; toothbrush; water cup; controller; towel; stapler;
whisk; towel from rod. \\
\addlinespace[2pt]
Place (6) & Apple into container; blue cube into plate; orange into small bowl;
marker pen into pen holder; red cube into paper cup; marker pen into open drawer. \\
\addlinespace[2pt]
Wipe (3) & Plate with rag; water stain with sponge; table with eraser. \\
\addlinespace[2pt]
Stack (7) & Red cube on yellow block; paper cup on paper cup; bowl on bowl;
coaster on coaster; cola can on cola can; cutlery-box lid; trash-can lid. \\
\addlinespace[2pt]
Pass (5) & Remote control, left-to-right; orange ball, right-to-left; sponge
eraser, left-to-right; Rubik's cube, right-to-left; cola can, left-to-right. \\
\addlinespace[2pt]
Fold (4) & Towel, one arm; towel, both arms; rag twice, both arms; unfold towel,
both arms. \\
\addlinespace[2pt]
Straighten (5) & Water bottle (two scene variants); water cup (two scene
variants); soda drink. \\
\addlinespace[2pt]
Pour (5) & Fruit pieces into plate; sesame into plate; sugar into glass cup;
beans into water cup; small cubes into bowl. \\
\addlinespace[2pt]
Push (8) & Red block left; paper cup to centre; black box left; kettle forward;
book to table edge; tissue pack into tissue box; close drawer; roll bottle. \\
\addlinespace[2pt]
Insert (6) & Bread into bread machine; flower into vase; straw into cup; plug
into power strip; remove book from shelf; place book into shelf. \\
\addlinespace[2pt]
Close (3) & Tissue-box lid; blue-box lid; kettle lid. \\
\addlinespace[2pt]
Separate (3) & Separate paper cups, both arms; uncap signature pen with the
opposite arm holding the pen; unscrew bottle cap, both arms. \\
\addlinespace[2pt]
Press (4) & Lamp switch; bread-machine button; stapler; calculator button. \\
\addlinespace[2pt]
Pull (6) & Remove pen cap; pull tissue from bag; pull tray forward; pull cable
from box; unplug charger while the opposite arm holds the power strip; open drawer. \\
\addlinespace[2pt]
Hang (4) & Remove towel from rod; hang towel on rod; hang rag on storage box,
left arm; hang hat on coat rack. \\
\addlinespace[2pt]
Tear (2) & Open envelope, left arm; tear one sheet from paper-towel roll, both arms. \\
\addlinespace[2pt]
Flip (6) & Close book; open book; turn book face-up; flip plate; flip coaster;
flip packaged tissues. \\
\addlinespace[2pt]
Sweep (3) & Sweep fruit cubes into dustpan with broom, both arms; sweep fruit
cubes by hand; sweep fruit cubes into dustpan with brush while the opposite arm
holds the dustpan. \\
\addlinespace[2pt]
Zip (3) & Zip file bag, both arms; unzip file bag, both arms; unzip pencil case
while the opposite arm holds the case. \\
\addlinespace[2pt]
Shake/Stir (4) & Shake glass bottle; shake bottle to check remaining amount;
stir coffee with chopstick; stir water with spoon. \\
\end{longtable}

\subsection{Task-Level Data-Scaling Results}

Table~\ref{tab:task_scaling_results} reports the empirical SR for every task,
data scale, and embodiment.  G1-OP and G2-90D results are shown in separate
four-column blocks, with every value computed over 10 trials.  The columns
correspond to the nested co-training pools used in Section~\ref{sec:data_scaling}.
The group headings and task descriptions match the taxonomy in
Table~\ref{tab:atomic_taxonomy}.

\newpage
\begingroup
\scriptsize
\setlength{\tabcolsep}{2pt}
\begin{longtable}{@{}p{1.70in}*{4}{>{\centering\arraybackslash}p{0.455in}}|*{4}{>{\centering\arraybackslash}p{0.455in}}@{}}
\caption{Task-level zero-shot OOD success rate (SR, \%) across data scales.
Every G1-OP and G2-90D result is shown separately; each SR is computed over
10 trials, and all 100 tasks are included.}\label{tab:task_scaling_results}\\
\toprule
Task & \multicolumn{4}{c|}{G1-OP SR (\%)} & \multicolumn{4}{c}{G2-90D SR (\%)} \\
\cmidrule(lr){2-5}\cmidrule(l){6-9}
& 0.3k h & 1.2k h & 5k h & 30k h & 0.3k h & 1.2k h & 5k h & 30k h \\
\midrule
\endfirsthead
\caption[]{Task-level zero-shot OOD success rate across data scales (continued).}\\
\toprule
Task & \multicolumn{4}{c|}{G1-OP SR (\%)} & \multicolumn{4}{c}{G2-90D SR (\%)} \\
\cmidrule(lr){2-5}\cmidrule(l){6-9}
& 0.3k h & 1.2k h & 5k h & 30k h & 0.3k h & 1.2k h & 5k h & 30k h \\
\midrule
\endhead
\bottomrule
\endfoot
\multicolumn{5}{@{}l|}{\textbf{Pick} (13 tasks)} & & & & \\
pick up red apple & 90 & 100 & 100 & 100 & 60 & 100 & 100 & 100 \\
pick up toothpaste & 0 & 20 & 40 & 80 & 30 & 40 & 60 & 60 \\
pick up red cube & 80 & 80 & 70 & 90 & 70 & 70 & 50 & 50 \\
pick up water bottle & 40 & 50 & 70 & 90 & 60 & 100 & 90 & 100 \\
pick up water bottle (2) & 50 & 70 & 90 & 100 & 100 & 100 & 100 & 100 \\
pick pen from water cup & 0 & 10 & 30 & 60 & 0 & 30 & 40 & 40 \\
pick up toothbrush & 30 & 30 & 40 & 50 & 0 & 40 & 30 & 30 \\
pick up water cup & 20 & 40 & 40 & 80 & 40 & 60 & 50 & 50 \\
pick up controller & 10 & 30 & 60 & 50 & 80 & 100 & 80 & 40 \\
pick up towel & 80 & 90 & 80 & 100 & 70 & 80 & 80 & 70 \\
pick up stapler & 70 & 100 & 80 & 90 & 80 & 80 & 70 & 80 \\
pick up whisk & 60 & 50 & 70 & 80 & 40 & 50 & 70 & 70 \\
pick towel from rod & 0 & 20 & 50 & 70 & 0 & 60 & 60 & 70 \\
\addlinespace[2pt]
\multicolumn{5}{@{}l|}{\textbf{Place} (6 tasks)} & & & & \\
put apple into container & 100 & 100 & 100 & 100 & 100 & 100 & 90 & 90 \\
put blue cube into plate & 100 & 100 & 100 & 100 & 90 & 90 & 100 & 100 \\
put orange into bowl & 80 & 100 & 100 & 100 & 60 & 60 & 80 & 80 \\
put pen into holder & 20 & 20 & 40 & 50 & 0 & 40 & 30 & 70 \\
put cube into paper cup & 30 & 60 & 80 & 80 & 30 & 0 & 40 & 60 \\
put pen into drawer & 60 & 60 & 90 & 100 & 70 & 50 & 70 & 80 \\
\addlinespace[2pt]
\multicolumn{5}{@{}l|}{\textbf{Wipe} (3 tasks)} & & & & \\
wipe plate with rag & 20 & 20 & 30 & 70 & 0 & 0 & 10 & 60 \\
wipe stain with sponge & 30 & 40 & 40 & 80 & 0 & 0 & 20 & 50 \\
wipe table with eraser & 40 & 40 & 60 & 80 & 0 & 0 & 30 & 60 \\
\addlinespace[2pt]
\multicolumn{5}{@{}l|}{\textbf{Stack} (7 tasks)} & & & & \\
stack cube on block & 0 & 10 & 20 & 40 & 0 & 0 & 0 & 10 \\
stack paper cups & 0 & 0 & 0 & 30 & 0 & 0 & 0 & 10 \\
stack bowls & 70 & 80 & 80 & 90 & 90 & 90 & 100 & 90 \\
stack coasters & 80 & 70 & 80 & 100 & 70 & 70 & 80 & 70 \\
stack cola cans & 0 & 0 & 0 & 10 & 0 & 0 & 0 & 10 \\
cover cutlery box lid & 0 & 0 & 0 & 0 & 0 & 0 & 0 & 0 \\
cover trash can lid & 0 & 0 & 0 & 20 & 0 & 0 & 0 & 0 \\
\addlinespace[2pt]
\multicolumn{5}{@{}l|}{\textbf{Pass} (5 tasks)} & & & & \\
pass remote control & 40 & 60 & 60 & 80 & 0 & 20 & 30 & 60 \\
pass orange ball & 70 & 70 & 80 & 100 & 0 & 0 & 0 & 0 \\
pass sponge eraser & 60 & 90 & 100 & 100 & 40 & 50 & 60 & 80 \\
pass rubik's cube & 0 & 0 & 0 & 0 & 0 & 0 & 0 & 0 \\
pass cola can & 0 & 40 & 60 & 70 & 0 & 20 & 20 & 30 \\
\addlinespace[2pt]
\multicolumn{5}{@{}l|}{\textbf{Fold} (4 tasks)} & & & & \\
fold towel (one arm) & 0 & 0 & 40 & 50 & 30 & 20 & 30 & 10 \\
fold towel (two arms) & 0 & 0 & 0 & 10 & 0 & 0 & 0 & 0 \\
fold rag twice & 0 & 0 & 0 & 0 & 0 & 0 & 0 & 0 \\
unfold folded towel & 20 & 0 & 0 & 0 & 0 & 0 & 0 & 0 \\
\addlinespace[2pt]
\multicolumn{5}{@{}l|}{\textbf{Straighten} (5 tasks)} & & & & \\
straighten water bottle & 0 & 0 & 0 & 10 & 0 & 0 & 0 & 10 \\
straighten water bottle (2) & 0 & 0 & 0 & 40 & 0 & 0 & 10 & 50 \\
straighten water cup & 0 & 0 & 0 & 60 & 0 & 0 & 0 & 40 \\
straighten water cup (2) & 0 & 0 & 0 & 30 & 0 & 0 & 0 & 0 \\
straighten soda drink & 0 & 0 & 0 & 10 & 0 & 0 & 0 & 20 \\
\addlinespace[2pt]
\multicolumn{5}{@{}l|}{\textbf{Pour} (5 tasks)} & & & & \\
pour fruits into plate & 20 & 40 & 50 & 70 & 30 & 20 & 20 & 30 \\
pour sesame into plate & 30 & 40 & 30 & 60 & 0 & 0 & 10 & 40 \\
pour sugar into cup & 10 & 20 & 30 & 50 & 0 & 0 & 0 & 10 \\
pour beans into cup & 0 & 20 & 40 & 50 & 0 & 0 & 10 & 10 \\
pour cubes into bowl & 0 & 0 & 0 & 40 & 0 & 30 & 0 & 10 \\
\addlinespace[2pt]
\multicolumn{5}{@{}l|}{\textbf{Push} (8 tasks)} & & & & \\
push block to left & 30 & 40 & 0 & 70 & 0 & 0 & 0 & 10 \\
push cup to center & 0 & 20 & 30 & 60 & 0 & 20 & 30 & 50 \\
push box to left & 40 & 40 & 60 & 70 & 0 & 0 & 10 & 20 \\
push kettle forward & 30 & 20 & 20 & 50 & 0 & 60 & 20 & 30 \\
push book to edge & 0 & 0 & 0 & 30 & 0 & 0 & 0 & 0 \\
push tissue pack in & 0 & 0 & 0 & 0 & 0 & 0 & 0 & 10 \\
close the drawer & 20 & 50 & 60 & 70 & 0 & 0 & 0 & 0 \\
roll the bottle & 0 & 0 & 0 & 10 & 0 & 0 & 10 & 10 \\
\addlinespace[2pt]
\multicolumn{5}{@{}l|}{\textbf{Insert} (6 tasks)} & & & & \\
insert bread into machine & 0 & 0 & 0 & 50 & 30 & 30 & 40 & 50 \\
insert flower into vase & 0 & 0 & 0 & 40 & 0 & 0 & 20 & 30 \\
insert straw into cup & 20 & 30 & 40 & 70 & 30 & 40 & 30 & 40 \\
insert plug into strip & 0 & 0 & 0 & 0 & 0 & 0 & 0 & 0 \\
take book from shelf & 0 & 0 & 0 & 60 & 0 & 0 & 20 & 50 \\
put book into shelf & 0 & 0 & 0 & 10 & 20 & 30 & 30 & 40 \\
\addlinespace[2pt]
\multicolumn{5}{@{}l|}{\textbf{Close} (3 tasks)} & & & & \\
close tissue box lid & 10 & 30 & 40 & 70 & 0 & 50 & 40 & 40 \\
close blue box lid & 30 & 50 & 40 & 90 & 0 & 70 & 60 & 80 \\
close kettle lid & 30 & 60 & 90 & 100 & 20 & 30 & 30 & 40 \\
\addlinespace[2pt]
\multicolumn{5}{@{}l|}{\textbf{Separate} (3 tasks)} & & & & \\
separate paper cups & 0 & 0 & 0 & 20 & 0 & 20 & 20 & 20 \\
uncap signature pen & 0 & 0 & 0 & 80 & 0 & 10 & 10 & 30 \\
unscrew bottle cap & 0 & 0 & 0 & 30 & 0 & 10 & 0 & 0 \\
\addlinespace[2pt]
\multicolumn{5}{@{}l|}{\textbf{Press} (4 tasks)} & & & & \\
press lamp switch & 0 & 0 & 0 & 0 & 0 & 0 & 0 & 0 \\
press bread machine button & 0 & 0 & 10 & 30 & 0 & 0 & 0 & 0 \\
press stapler to staple & 0 & 0 & 10 & 30 & 0 & 0 & 10 & 20 \\
press calculator button & 0 & 0 & 0 & 10 & 0 & 0 & 0 & 10 \\
\addlinespace[2pt]
\multicolumn{5}{@{}l|}{\textbf{Pull} (6 tasks)} & & & & \\
pull off pen cap & 0 & 0 & 0 & 20 & 0 & 0 & 0 & 20 \\
pull tissue from bag & 10 & 30 & 30 & 80 & 0 & 0 & 30 & 40 \\
pull tray to front & 0 & 0 & 10 & 30 & 0 & 0 & 0 & 10 \\
pull cable from box & 0 & 0 & 0 & 0 & 0 & 0 & 0 & 0 \\
unplug charger from strip & 0 & 0 & 0 & 0 & 0 & 0 & 0 & 10 \\
open the drawer & 0 & 0 & 0 & 10 & 0 & 0 & 0 & 0 \\
\addlinespace[2pt]
\multicolumn{5}{@{}l|}{\textbf{Hang} (4 tasks)} & & & & \\
take towel off rod & 0 & 0 & 10 & 60 & 0 & 0 & 30 & 50 \\
hang towel on rod & 0 & 0 & 0 & 0 & 0 & 0 & 0 & 0 \\
hang rag on box & 20 & 0 & 0 & 0 & 0 & 30 & 20 & 20 \\
hang hat on rack & 0 & 0 & 0 & 0 & 0 & 0 & 0 & 0 \\
\addlinespace[2pt]
\multicolumn{5}{@{}l|}{\textbf{Tear} (2 tasks)} & & & & \\
tear open envelope & 0 & 0 & 0 & 10 & 0 & 0 & 0 & 10 \\
tear paper towel off roll & 0 & 0 & 0 & 0 & 0 & 0 & 0 & 0 \\
\addlinespace[2pt]
\multicolumn{5}{@{}l|}{\textbf{Flip} (6 tasks)} & & & & \\
close the open book & 0 & 50 & 70 & 90 & 0 & 0 & 20 & 30 \\
open the book & 0 & 0 & 0 & 0 & 0 & 0 & 0 & 0 \\
flip book to front & 0 & 0 & 0 & 0 & 0 & 0 & 0 & 0 \\
flip plate over & 0 & 0 & 0 & 0 & 0 & 0 & 0 & 0 \\
flip coaster to front & 60 & 70 & 60 & 80 & 0 & 60 & 70 & 90 \\
flip tissue pack over & 0 & 0 & 0 & 0 & 0 & 0 & 0 & 0 \\
\addlinespace[2pt]
\multicolumn{5}{@{}l|}{\textbf{Sweep} (3 tasks)} & & & & \\
sweep cubes into dustpan & 0 & 0 & 0 & 10 & 0 & 0 & 0 & 0 \\
sweep cubes by hand & 0 & 0 & 0 & 0 & 0 & 0 & 0 & 10 \\
sweep cubes with brush & 0 & 0 & 0 & 0 & 0 & 0 & 0 & 10 \\
\addlinespace[2pt]
\multicolumn{5}{@{}l|}{\textbf{Zip} (3 tasks)} & & & & \\
zip up file bag & 0 & 0 & 0 & 0 & 0 & 0 & 0 & 0 \\
unzip file bag & 0 & 0 & 0 & 0 & 0 & 0 & 0 & 0 \\
unzip pencil case & 0 & 0 & 0 & 0 & 0 & 0 & 0 & 0 \\
\addlinespace[2pt]
\multicolumn{5}{@{}l|}{\textbf{Shake/Stir} (4 tasks)} & & & & \\
shake glass bottle & 0 & 0 & 10 & 20 & 0 & 40 & 40 & 60 \\
shake bottle to check & 0 & 0 & 10 & 30 & 0 & 30 & 40 & 60 \\
stir coffee with chopstick & 0 & 0 & 0 & 0 & 0 & 0 & 0 & 0 \\
stir water with spoon & 0 & 0 & 0 & 0 & 0 & 0 & 0 & 10 \\
\addlinespace[2pt]
\end{longtable}
\endgroup

\subsection{Instruction-Following Prompts and Results}
\label{app:language_details}

\paragraph{Protocol.}
The fine-grained grounding study of Section~\ref{sec:language_dimensions}
evaluates the final checkpoint of the full \methodname{} co-training run on
G1-OP without any adaptation.  The suite contains 59 instructions, 33 for
Pick and 26 for Place, and every instruction is executed for five trials
under the same reset procedure, giving 295 rollouts.  A trial satisfies the
Follow Score when the gripper contacts only the instruction-specified object
(Pick) or the carried object contacts only the specified destination
container (Place); SR additionally requires the complete Pick or Place
behavior.  The instruction-level Follow Score and SR reported in
Table~\ref{tab:if_instructions} are the fraction of the five trials that
satisfy each criterion, so every value is a multiple of 20\%.  Averaged over
all 295 rollouts, the Follow Score is 83.1\% and the SR is 72.9\%.

\paragraph{Scenes.}
Each of the five physical scene families in Figure~\ref{fig:language_scenes}
has a Pick and a Place variant, and each is arranged so that several actions
are visually feasible while only the instruction-specified target or
destination is correct.
(i)~\emph{Object identity}: a cluttered tabletop of distinct household items
(a badminton ball, a knife, a comb, a computer mouse, a television remote, a
toy bear, a screwdriver, and a roll of tape) for Pick, and a white plate, a
box, a bowl, and a cup as candidate destinations for Place.
(ii)~\emph{Color}: fruits in four colors (red, orange, yellow, and green) for
Pick, and plates in four colors (blue, pink, orange, and yellow) for Place.
(iii)~\emph{Size}: blocks or cups in three sizes for Pick and containers in
three sizes for Place, each evaluated in two layouts; the instructions use
both extreme references (largest, smallest, larger) and the medium reference.
(iv)~\emph{Shape}: rectangular, cylindrical, and round objects for Pick and
containers of the same three shapes for Place.
(v)~\emph{Position and order}: a row of identical cups (Pick) or containers
(Place).  The horizontal layout supports left/right references and the
set-relative references \emph{middle} and \emph{second from the left/right};
the vertical layout supports near-side and far-end references.

\paragraph{Dimension mapping.}
Each instruction is annotated with one raw subcategory, which maps to the six
grounding dimensions as follows.  Object identity and color are single
subcategories.  Size combines extreme references (largest, smallest, larger)
and the medium reference.  Position combines direct left/right references and
near/far references.  Order combines the middle reference and ordinal
references (second from the left or right).  Table~\ref{tab:if_instructions}
lists the subcategory beside every instruction.  Instructions are reproduced
verbatim as issued to the policy, including minor typographical
irregularities.

\begingroup
\small
\setlength{\tabcolsep}{4pt}
\begin{longtable}{@{}p{1.15in}p{3.55in}rr@{}}
\caption{Instruction-level results of the fine-grained grounding study
(G1-OP, five trials per instruction).  Follow and SR are percentages of the
five trials.}\label{tab:if_instructions}\\
\toprule
Dimension & Instruction & Follow & SR \\
\midrule
\endfirsthead
\caption[]{Instruction-level results of the fine-grained grounding study (continued).}\\
\toprule
Dimension & Instruction & Follow & SR \\
\midrule
\endhead
\bottomrule
\endfoot
\multicolumn{4}{@{}l}{\textbf{Pick} (33 instructions)} \\
Object & pick up the white badminton ball with the right arm. & 100 & 80 \\
Object & pick up the yellow knife with the left arm. & 100 & 80 \\
Object & pick up the blue comb with the right arm. & 100 & 80 \\
Object & pick up the black mouse with the left arm. & 100 & 100 \\
Object & pick up the black tv controller with the right arm. & 100 & 100 \\
Object & pick up the toy bear with the left arm. & 100 & 60 \\
Object & pick up the yellow screwdriver with the right arm. & 100 & 80 \\
Object & pick up the green tape with the right arm. & 100 & 100 \\
Color & pick up the red fruit with the right arm. & 100 & 60 \\
Color & pick up the orange fruit with the right arm. & 100 & 100 \\
Color & pick up the yellow fruit with the right arm. & 100 & 100 \\
Color & pick up the green fruit with the right arm. & 100 & 100 \\
Size (extreme) & pick up the largest block with the right arm. & 100 & 100 \\
Size (extreme) & pick up the smallest block with the left arm. & 100 & 100 \\
Size (medium) & pick up the medium block with the right arm. & 0 & 0 \\
Size (extreme) & pick up the larger cup with the left arm. & 100 & 80 \\
Size (medium) & pick up the medium block with the left arm. & 100 & 100 \\
Size (extreme) & pick up the largest block with the right arm. & 100 & 100 \\
Size (extreme) & pick up the smallest block with the right arm. & 100 & 80 \\
Size (medium) & pick up the medium block with the right arm. & 60 & 40 \\
Shape & pick up the rectangular object with the right arm. & 100 & 20 \\
Shape & pick up the cylindrical object with right arm. & 60 & 60 \\
Shape & pick up the round object with right arm. & 100 & 100 \\
Shape & pick up the cylindrical object with left arm. & 100 & 100 \\
Position (left/right) & pick up the cup on the right side. & 80 & 20 \\
Order (ordinal) & Pick up the second cup from the right. & 0 & 0 \\
Order (middle) & Pick up the middle cup. & 40 & 40 \\
Order (ordinal) & Pick up the second cup from the left. & 0 & 0 \\
Position (left/right) & pick up the cup on the left side. & 100 & 100 \\
Position (left/right) & pick up the cup on the right of the table with right arm. & 100 & 100 \\
Position (left/right) & pick up the cup on the left of the table with left arm. & 100 & 100 \\
Position (near) & Pick up the cup on the near side of the table with the right arm. & 60 & 40 \\
Position (far) & pick up the farthest cup on the table with right arm. & 100 & 100 \\
\addlinespace[3pt]
\multicolumn{4}{@{}l}{\textbf{Place} (26 instructions)} \\
Object & place the object into the plate with the right arm. & 100 & 100 \\
Object & place the object into the box with the right arm. & 100 & 100 \\
Object & place the object into the bowl with the right arm. & 100 & 80 \\
Object & place the object into the cup with the right arm. & 100 & 60 \\
Color & place the object into the blue plate with the right arm. & 100 & 100 \\
Color & place the object into the pink plate with the right arm. & 100 & 100 \\
Color & place the object into the orange plate with the right arm. & 100 & 100 \\
Color & place the object into the yellow plate with the right arm. & 100 & 100 \\
Size (extreme) & place the object into the smallest container with left arm. & 100 & 40 \\
Size (extreme) & place the object into the largest container with right arm. & 100 & 100 \\
Size (medium) & place the object into the medium container with right arm. & 0 & 0 \\
Size (medium) & place the object into the medium container with left arm. & 80 & 80 \\
Size (extreme) & place the object into the smallest container with right arm & 0 & 0 \\
Size (extreme) & place the object into the largest container with right arm. & 80 & 80 \\
Size (medium) & place the object into the medium container with right arm. & 100 & 100 \\
Shape & place the object into the rectangular container with the right arm. & 100 & 100 \\
Shape & place the objectinto the cylindrical container with the right arm & 80 & 40 \\
Shape & place the object into the round container with the right arm & 80 & 60 \\
Shape & place the object into the round container with the left arm & 100 & 100 \\
Position (left/right) & place the object into the container on the right side with the right arm. & 100 & 100 \\
Order (middle) & place the object into the middle container with the right arm. & 20 & 20 \\
Order (middle) & place the object into the middle container with the left arm. & 40 & 40 \\
Position (left/right) & place the object into the container on the left side with the left arm. & 100 & 100 \\
Position (far) & Place the object into the container at the far end of the table with the right arm. & 100 & 100 \\
Position (near) & Place the object into the container on the near side of the table with the right arm. & 100 & 60 \\
Order (ordinal) & place the object into the second container from the left. & 20 & 20 \\
\end{longtable}
\endgroup

\paragraph{Descriptor frequencies.}
The occurrence rates in Figure~\ref{fig:language_following} are lexical
matches over two corpora: the 95{,}010 human-written referring expressions of
RefCOCOg~\citep{mao2016generation}, and the step-captioned Pick and Place
instruction segments of the GE-Act~2.0 training corpus (approximately 3.7
million segments, simulation excluded).  Table~\ref{tab:if_lexical} gives the
matching rule for every dimension together with the RefCOCOg counts and the
resulting shares.  Non-object descriptors are counted independently, so one
expression can contribute to several dimensions and the shares are not a
partition; object identity is defined as the residual expressions that match
none of the other five rules in RefCOCOg, and as bare object noun phrases
without a descriptor in the training corpus.  Size is the one dimension for
which simulation instructions are included in the training-corpus share,
because superlative size references occur almost exclusively there; its
0.96\% comprises 0.68\% extreme and 0.28\% medium references.

\begin{table}[H]
\captionsetup{skip=7pt}
\caption{Lexical matching rules and descriptor frequencies used in
Figure~\ref{fig:language_following}.  RefCOCOg counts are out of 95{,}010
expressions.}
\label{tab:if_lexical}
\centering
\tablebodyfont
\setlength{\tabcolsep}{5pt}
\begin{tabular}{@{}lp{2.55in}rrr@{}}
\toprule
Dimension & Matching rule & RefCOCOg count & RefCOCOg (\%) & GE corpus (\%) \\
\midrule
Object & Residual: no color, size, shape, position, or order descriptor & 39{,}474 & 41.5 & 58.3 \\
Color & Explicit color words; \emph{orange} as a fruit name excluded & 42{,}161 & 44.4 & 29.6 \\
Size & \emph{large}, \emph{small}, \emph{big}, \emph{tiny}, \emph{medium} and comparative/superlative forms & 5{,}299 & 5.6 & 0.96 \\
Shape & \emph{round}, \emph{square}, \emph{rectangular}, \emph{cylindrical}, \emph{triangular} & 574 & 0.6 & 0.67 \\
Position & \emph{left}, \emph{right}, \emph{near}, \emph{far}; body-side phrases such as \emph{right hand} excluded unless another spatial relation is present & 11{,}927 & 12.6 & 13.8 \\
Order & \emph{middle}, \emph{center}, ordinals, \emph{leftmost}, \emph{rightmost}; generic \emph{between} excluded & 2{,}440 & 2.6 & 0.13 \\
\bottomrule
\end{tabular}
\end{table}


\end{document}